\documentclass[journal]{IEEEtran}
\usepackage{amsmath,amsfonts}
\usepackage{algorithmic}
\usepackage{algorithm}
\usepackage{array}
\usepackage{fontawesome5}
\usepackage[colorlinks=true, linkcolor=red, citecolor=ngreen, urlcolor=blue]{hyperref}
\usepackage[caption=false,font=normalsize,labelfont=sf,textfont=sf]{subfig}
\usepackage{textcomp}
\usepackage{multirow}
\usepackage{stfloats}
\usepackage{url}
\usepackage{flushend}
\usepackage{verbatim}
\usepackage{graphicx}
\usepackage{relsize}
\usepackage{pgfplots}
\usepackage{colortbl}
\usepackage{cite}
\usepackage{float}
\usepackage{dsfont}
\usepackage{orcidlink}
\usepackage{arydshln}
\usepackage{enumitem}
\usepackage{float}
\usepackage{hyperref}
\usepackage{cleveref}
\usepackage{multirow}
\usepackage{amssymb}
\usepackage{booktabs}
\usepackage{bbm}
\usepackage{tikz}
\usepackage{bm}
\usepackage[normalem]{ulem}
\usepackage{xcolor}
\usepackage{xspace}  

\pgfplotsset{compat=1.18}

\begin{document}
\definecolor{ngreen}{RGB}{17, 173, 30}
\newcommand{\reddashedline}{\textcolor{red}{\rule[0.5ex]{0.3em}{1pt}\hspace{0.2em}\rule[0.5ex]{0.3em}{1pt}\hspace{0.2em}\rule[0.5ex]{0.3em}{1pt}\hspace{0.2em}\rule[0.5ex]{0.3em}{1pt}}}
\definecolor{mycolor_green}{RGB}{88, 142, 49}
\definecolor{mycolor_red}{RGB}{192, 0, 0}
\definecolor{mycolor_orange}{RGB}{242, 186, 2}
\newcommand{\cyh}[1]{{\color{red}{[YH: #1]}}}
\newcommand{\tychen}[1]{{\color{orange}{[tychen: #1]}}}
\newcommand{\hbliu}[1]{{\color{blue}{[hbliu: #1]}}}
\newcommand{\mycite}[1]{\textcolor{red}{#1}}
\setlength{\arrayrulewidth}{1mm}
\title{Awakening Diffusion Transformers: Eliciting Stronger Generation and Understanding via Massive Activation Modulation}

\author{Chaofan Gan, Zicheng Zhao, Yuanpeng Tu, Xi Chen, Ziran Qin, Tieyuan Chen, Supavadee Aramvith, \\Junhui Hou,~\IEEEmembership{Senior Member, IEEE}, Mehrtash Harandi,~\IEEEmembership{Member, IEEE}, Weiyao Lin,~\IEEEmembership{Senior Member, IEEE}
\thanks{The paper is supported in part by the National Natural Science Foundation of China (No. 62595733, 62325109, 62561160155), in part by the Hong Kong Research Grants Council under Grant N\_CityU1114/25, and in part by the Shanghai 'The Belt and Road' Young Scholar Exchange Grant (24510742000).
\textit{(Corresponding authors: Weiyao Lin.)}}
\thanks{Chaofan Gan, Zicheng Zhao, Ziran Qin, Tieyuan Chen, Weiyao Lin are with Shanghai Jiao Tong University, Shanghai, China. Mehrtash Harandi is with Monash University, Melbourne, Australia. Yuanpeng Tu and Xi Chen are with The University of Hong Kong, Hong Kong, China. Chaofan Gan is also with Monash University, Melbourne, Australia. Tieyuan Chen and Weiyao Lin are also with the Zhongguancun Academy, Beijing, China. Supavadee Aramvith is with Chulalongkorn University, Bangkok, Thailand. Junhui Hou is with City University of Hong Kong, Hong Kong, China. E-mail:\{ganchaofan, wylin\}@sjtu.edu.cn.}
}
\markboth{}%
{Gan \MakeLowercase{\textit{et al.}}: Massive Activations in DiTs}

\maketitle

\begin{figure*}[htp]
  \centering
  \includegraphics[width =\linewidth]{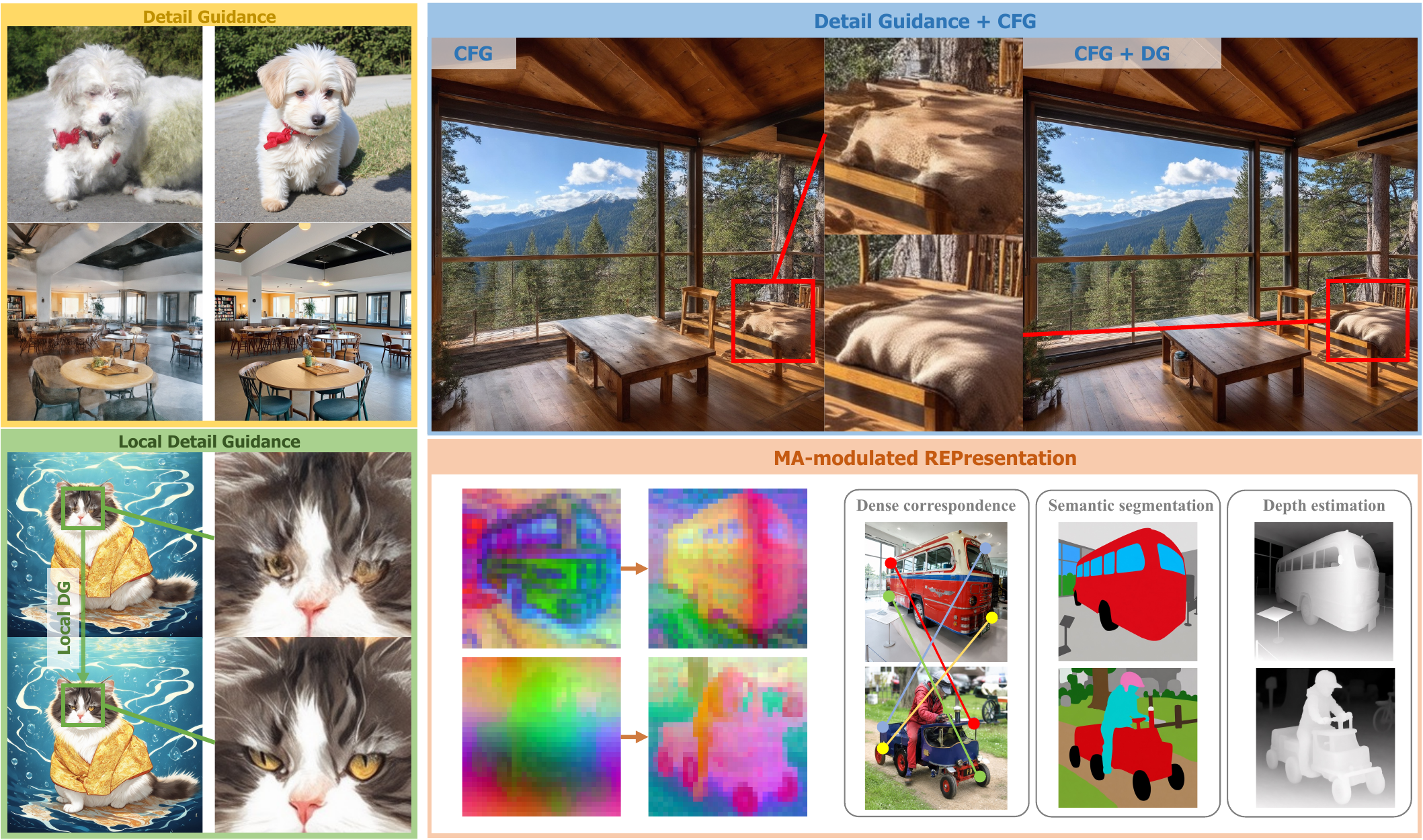}
  \vspace{-2mm}
\caption{\textbf{EMA improves DiTs across visual generation and understanding tasks.}
  For generation, EMA improves fine-grained detail synthesis and enables controllable local refinement. 
  For understanding, EMA produces more discriminative DiT representations, 
  benefiting dense perception tasks such as correspondence and segmentation.}

    \vspace{-1mm}
  \label{figure:f1}
\end{figure*}

\begin{abstract}
Massive Activations (MAs) have been widely observed in Transformer-based models, yet
their structure and functional roles in Diffusion Transformers (DiTs) remain
insufficiently understood. In this work, we systematically analyze MAs in
representative DiTs and find that they are spatially distributed across image tokens
while concentrated in a small set of fixed feature dimensions. We further show that
these dimensions are closely aligned with AdaLN residual scaling factors and are
primarily modulated by the denoising timestep rather than text conditions. This
structure leads to two task-dependent effects: for generation, MAs are critical for
fine-grained detail synthesis while having limited influence on global semantics; for
understanding, their shared high-magnitude directions make raw DiT features overly
similar across spatial tokens and weaken dense feature discrimination.
Based on these findings, we introduce \textbf{E}liciting \textbf{M}assive
\textbf{A}ctivation (\textbf{EMA}), a training-free framework that leverages Massive Activations (MAs) as a unified modulation signal to
improve both generative and representational capabilities of DiTs. For generation,
EMA proposes MA-driven \textbf{D}etail \textbf{G}uidance (\textbf{DG}), which
suppresses MA dimensions to construct a detail-deficient counterfactual prediction
and guides sampling toward finer visual details. DG further supports efficient
partial-forward inference, integration with classifier-free guidance, and token-level
Local DG for refining selected image regions. For understanding, EMA introduces
\textbf{M}A-modulated \textbf{REP}resentation extraction (\textbf{MREP}), which uses
pretrained AdaLN channel-wise modulation to reduce MA directional dominance and
concatenates spatially normalized MA maps to preserve useful spatial structure.
Extensive experiments on image generation, video generation, local detail refinement,
and dense visual understanding demonstrate that EMA consistently improves both the
generation quality and representation capability of DiTs.
\end{abstract}


\section{Introduction}

\IEEEPARstart{D}{iffusion} models~\cite{rombach2022high,saharia2022photorealistic}
have become a dominant paradigm for high-quality visual generation. With the adoption
of Transformer denoisers~\cite{vaswani2017attention,peebles2023scalable}, Diffusion
Transformers (DiTs) now underpin many strong text-to-image and text-to-video systems,
including SD3, Flux, and recent video DiTs~\cite{esser2024scaling,Flux,yang2024cogvideox,hong2022cogvideo,wan2025}.
At the same time, pretrained diffusion models are increasingly reused as visual
feature extractors for dense correspondence and perceptual
tasks~\cite{hedlin2024unsupervised,tang2023emergent,zhang2024tale,li2024sd4match}.
These two trends place DiTs in a dual role: they are powerful generators, and they
also contain representations that may support visual understanding. However, the
internal activation components that determine these two capabilities remain
insufficiently understood.

A particularly important but underexplored component is \emph{Massive Activations}
(MAs), i.e., hidden activations whose magnitudes are much larger than the remaining
activations. MAs have been studied in large language models, where they are related
to contextual knowledge and long-context modeling~\cite{sun2024massive,xiao2024efficient},
and in Vision Transformers, where similar activation patterns are associated with
background or register-like tokens and global semantic processing~\cite{darcet2023vision}.
DiTs differ from both settings. They perform iterative denoising over a spatial grid,
where every image token participates in reconstructing local content and the hidden
state evolves with the denoising timestep. Therefore, MAs in DiTs cannot be fully
understood as token-localized language outliers or recognition-specific global
tokens. Their role must be examined in the context of dense visual generation.

\begin{figure*}[htp]
  \centering
  \includegraphics[width =\linewidth]{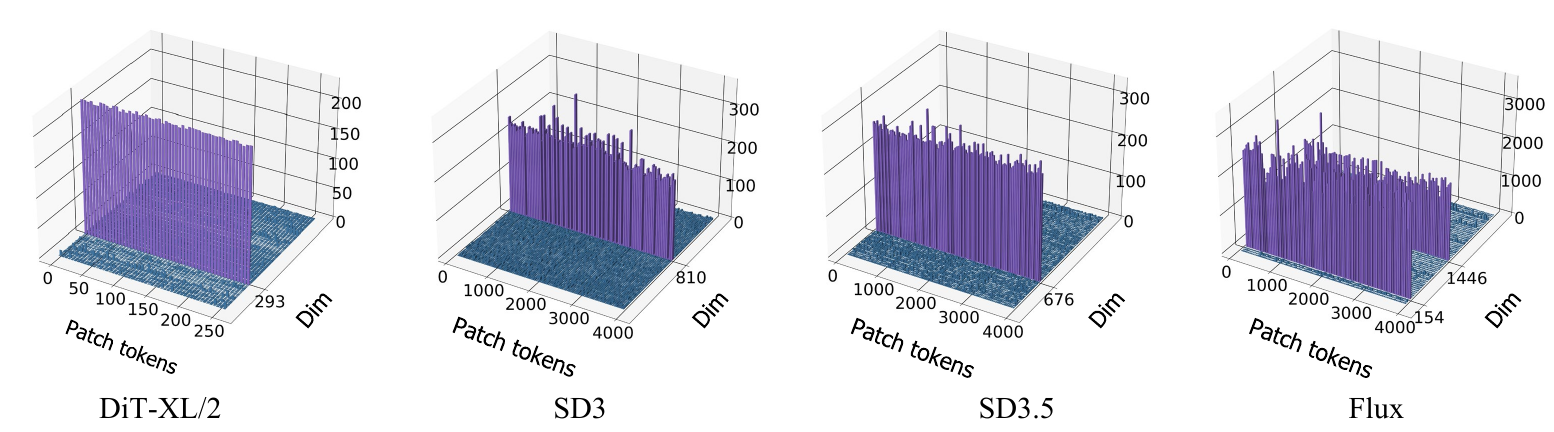}
  \caption{\textbf{Massive Activations in DiTs.} The activation magnitudes of internal
  hidden states. We present the average magnitudes over 1{,}000 text prompts. Massive
  activations are consistently concentrated in a few fixed dimensions across all
  patch tokens. }
  \vspace{-8mm}
  \label{figure:f2}
\end{figure*}

In this work, we systematically analyze MAs in representative DiTs. We find that MAs
exhibit a DiT-specific structure: they are distributed across spatial tokens, yet
concentrated in a small set of fixed feature dimensions (\Cref{figure:f2}). This
spatially distributed and channel-concentrated pattern suggests that MAs act as
shared high-magnitude channels for dense visual computation, rather than as isolated
outliers at a few special positions. We further show that the MA dimensions are
closely aligned with the residual scaling factors produced by Adaptive Layer
Normalization (AdaLN) (\Cref{figure:f4}). Their magnitudes are primarily shaped by the denoising
timestep, while text conditions have much weaker influence. These observations
indicate that MAs are structured, AdaLN-related, and stage-dependent activation
components inside pretrained DiTs.

This structure leads to different consequences for generation and understanding. For
visual generation, activation intervention shows that disrupting MA dimensions
substantially weakens fine-grained local details, such as textures and subtle object
parts, while largely preserving object identity, layout, and global semantic content.
In contrast, perturbing the same number of non-MA dimensions has much weaker visual
effect. This indicates that MAs are not the primary carrier of prompt semantics;
instead, they are closely tied to local detail synthesis. Thus, suppressing MAs gives
a natural way to construct a detail-deficient counterfactual branch whose residual
from the original prediction isolates a detail-specific guidance direction.

For visual understanding, the same MA structure creates a different problem. Dense
matching and perception require spatial descriptors whose directions reflect local
semantic and geometric differences. However, since MAs occupy fixed high-magnitude
dimensions across many spatial tokens, raw DiT features become biased toward a shared
dominant direction, making local descriptors overly similar under cosine-based
comparison. This weakens spatial discrimination even though the underlying DiT is a
strong generative backbone. Importantly, MA dimensions should not simply be discarded:
their spatial maps still exhibit meaningful response patterns over image tokens. A
useful representation should therefore reduce the directional dominance of MAs while
preserving their spatial structure.

Motivated by these findings, we introduce 
\textbf{E}liciting \textbf{M}assive \textbf{A}ctivation (\textbf{EMA}), 
a unified training-free framework that modulates MAs to improve both generation 
and understanding in DiTs, as shown in~\Cref{figure:f1}. For generation, 
EMA introduces MA-driven \textbf{D}etail \textbf{G}uidance (\textbf{DG}), 
which enhances fine-grained rendering by contrasting the original prediction 
with a detail-deficient counterfactual branch. Specifically, DG suppresses 
MA dimensions at an intermediate DiT block to obtain the detail-deficient 
prediction, and then guides the original prediction away from this branch toward 
more detail-rich outputs.
This hidden-state intervention naturally supports efficient partial-forward inference: the original and counterfactual branches share the computation before the intervention block, so only the remaining blocks need to be recomputed after MA suppression. Moreover, applying MA suppression only to selected spatial tokens yields Local DG, enabling targeted refinement of regions with weak details. DG is complementary to Classifier-Free Guidance (CFG)~\cite{ho2022classifier}: while CFG primarily strengthens semantic alignment with conditioning signals, DG explicitly improves fine-grained visual detail quality.
For understanding, EMA introduces \textbf{M}A-modulated \textbf{REP}resentation
extraction (\textbf{MREP}). MREP uses pretrained AdaLN channel-wise modulation to
reduce MA directional dominance, and concatenates spatially normalized MA maps to
retain their useful spatial cues. In this way, EMA treats MAs in a task-aware manner:
their suppression defines detail guidance for generation, while their modulated and
spatially normalized form improves dense representations.
Our main contributions are summarized as follows:
\begin{itemize}
\item
We provide a systematic study of MAs in DiTs and show that they are spatially
distributed, channel-concentrated, AdaLN-related, and primarily timestep-modulated
activation components.
\item
We reveal the dual role of MAs: they support fine-grained local detail synthesis
during generation, but their shared high-magnitude directions impair the
discriminability of raw DiT features for dense understanding.
\item
We introduce EMA, a training-free MA modulation framework with two task-specific
modules: DG for detail-faithful image and video generation, and MREP for
semantically discriminative DiT representation extraction.
\item
We demonstrate that EMA improves image generation, video generation, local detail
refinement, and dense visual understanding, establishing MA modulation as a unified
tool for eliciting stronger DiT capabilities.
\end{itemize}

\section{Related Work}

\subsection{Diffusion Transformers for Visual Generation}
Diffusion models~\cite{ho2020denoising, dhariwal2021diffusion, rombach2022high, saharia2022photorealistic} have become a 
dominant paradigm for high-quality visual synthesis. In particular, latent diffusion models~\cite{rombach2022high} improve 
computational efficiency by performing denoising in a compressed latent space, while large-scale text-to-image systems further 
demonstrate the importance of scaling model capacity, data, and conditioning signals~\cite{saharia2022photorealistic}.
Recently, Transformer architectures~\cite{vaswani2017attention, dosovitskiy2020image} have been increasingly adopted as 
denoising backbones. DiT~\cite{peebles2023scalable} introduces a pure Transformer formulation for diffusion models, and a 
series of DiT-based systems, including PixArt-$\alpha$~\cite{chen2023pixart}, SD3~\cite{esser2024scaling}, and Flux~\cite{Flux}, 
have demonstrated strong performance in text-to-image generation. Beyond image synthesis, Transformer-based diffusion models have 
also become central to text-to-video generation, such as CogVideo~\cite{hong2022cogvideo}, CogVideoX~\cite{yang2024cogvideox}, and 
Wan~\cite{wan2025}.
Despite their strong empirical success, most prior work primarily focuses on architecture design, training strategies, and scaling laws. 
In contrast, our work shifts the focus to the internal computations of DiT hidden states, aiming to better understand their role in 
enabling both visual generation and representation learning.

\subsection{Sampling Guidance for Diffusion Models}
Sampling guidance provides a practical mechanism for steering pretrained diffusion models without retraining the generator. 
Among existing methods, classifier-free guidance (CFG)~\cite{ho2022classifier} has become the de facto standard. It linearly extrapolates 
between unconditional and conditional predictions, thereby strengthening conditioning signals and improving prompt or class alignment. 
However, large guidance scales can lead to over-saturation and visual artifacts~\cite{sadat2024eliminating}, which motivates improved 
variants such as CFG++~\cite{chung2024cfg++}.
Beyond CFG-style interpolation, another line of work constructs guidance signals from internal model behavior. 
Auto-guidance~\cite{karras2024guiding} steers sampling using a deliberately degraded version of the same model, while self-attention 
guidance~\cite{hong2023improving}, perturbed-attention guidance~\cite{ahn2024self}, and spatiotemporal 
skip guidance~\cite{hyung2025spatiotemporal} modify attention maps or intermediate activations to obtain improved sampling directions.
Our method follows the general philosophy of self-guidance, but differs in both motivation and construction. 

\subsection{Diffusion Model Representations}
Beyond image synthesis, pretrained diffusion models have demonstrated strong potential as visual representation learners. 
During large-scale generative training, these models implicitly acquire spatial structure and compositional semantics, 
making their intermediate features useful for a range of downstream perception 
tasks~\cite{amit2021segdiff, baranchuk2021label, xu2023open, zhao2023unleashing}.
In visual correspondence, a line of work explores feature extraction from pretrained Stable Diffusion (SD) models for 
semantic or geometric matching across images~\cite{hedlin2024unsupervised, tang2023emergent, luo2024diffusion, zhang2024tale, li2024sd4match}. 
These approaches are often evaluated in comparison or combination with discriminative representations such as 
DINO and DINOv2~\cite{amir2021deep, oquab2023dinov2}, while recent studies further enhance geometric consistency via 
adaptive pose alignment~\cite{zhang2024telling}.
More recently, several works have begun to repurpose Diffusion Transformers (DiTs) as general-purpose feature extractors for 
correspondence and perception tasks~\cite{gan2025unleashing, son2025repurposing}, suggesting that stronger generative backbones may 
also yield more expressive visual representations.
However, despite these encouraging results, the internal representational properties of DiTs remain significantly less understood. 

\subsection{Massive Activations and Internal Outliers}

Massive activations (MAs) refer to unusually large hidden activations that concentrate in a small subset of feature dimensions or tokens. 
They were first systematically studied in large language models, where they often emerge at fixed dimensions of low-information tokens and 
have been linked to contextual knowledge modeling and long-context 
capabilities~\cite{sun2024massive, xiao2024efficient, zhao2023unveiling, xu2025slmrec}. Subsequent studies further associate such activation 
concentration with positional mechanisms such as rotary position embeddings~\cite{jin2025massive}.
Similar phenomena have also been observed in Vision Transformers~\cite{darcet2023vision, yang2024denoising, sun2024massive}, where 
high-magnitude activations are frequently associated with background or register-like tokens that help organize global semantic structure. 
Beyond discriminative models, massive or outlier activations have also been reported in Diffusion Transformers (DiTs), particularly in the 
context of quantization, acceleration, and feature extraction~\cite{liu2024hq, zhao2024vidit, fang2025tinyfusion, gan2025unleashing}.
Notably, DiTF~\cite{gan2025unleashing} observes that DiTs exhibit fixed-dimensional massive activations across spatial tokens when used as 
feature extractors, while DG~\cite{gan2025massive} further suggests that such activations are closely related to local detail synthesis in 
DiT-based generation. However, despite these empirical observations, the role of MAs during the denoising process and representation 
formation in DiTs remains poorly understood.

\begin{figure*}[tp]
  \centering
  \includegraphics[width =\linewidth]{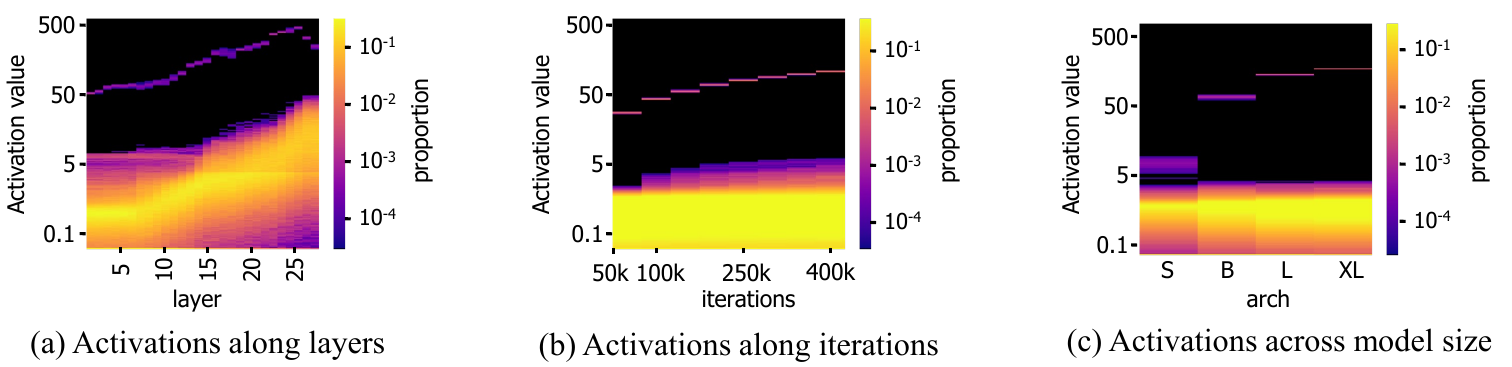}
  \caption{\textbf{Properties of massive activations in
  DiT-XL.} (a) Activation distribution of the hidden states along DiT layers (b)
  Activation distribution of the hidden states along training iterations (c) Activation
  distribution of the hidden states across different model sizes. Massive activations
  occur throughout all layers and persist across different model sizes.}
  \vspace{-2mm}
  \label{figure:f3}
\end{figure*}

\begin{figure*}[tp]
  \centering
  \includegraphics[width =\linewidth]{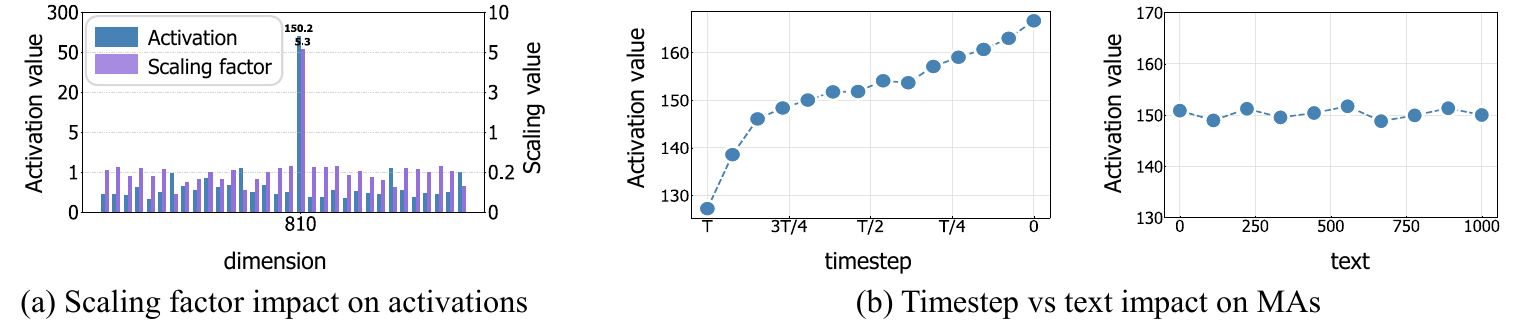}
  \caption{\textbf{Impact of the input timestep and text on Massive Activations (MAs).} 
  (a) Comparison of the distributions of hidden-state $z_t^k$ activations and
  their corresponding residual scaling factor $\alpha_k$. (b) Respective impact of
  input timestep and text embeddings on the magnitude distribution of MAs, where we
  compare the MAs of 1000 different text inputs. The massive activations are governed
  by the residual scaling factor; their magnitude is primarily shaped by the input
  timestep embedding $t$, while text embeddings $c$ have negligible effect.}
  \vspace{-5mm}
  \label{figure:f4}
\end{figure*}

\section{Preliminaries}
\label{preliminary}
\subsection{Diffusion Transformer Architecture}
Following DiT~\cite{peebles2023scalable}, we consider a latent diffusion model composed
of a variational autoencoder (VAE) and a denoising Transformer. The VAE encoder
$\mathcal{E}$ maps an image $x$ into a latent variable $z_0=\mathcal{E}(x)$, and the VAE
decoder maps the denoised latent back to pixel space. The Transformer denoiser
$D_\theta$ operates on a noised latent $z_t \in \mathbb{R}^{C \times H \times W}$ at
timestep $t$ under condition $c$. Let $D_\theta=\{D_k\}_{k=1}^{N}$ contain $N$
Transformer blocks, and denote the hidden state after the $k$-th block as $z_t^k$.
Each block updates the hidden state through a residual connection~\cite{he2016deep}:
\begin{equation}
z_t^k=z_t^{k-1}+\alpha_{k-1} D_{k-1}(z_t^{k-1},t,c),
\label{block}
\end{equation}
where $\alpha_k \in \mathbb{R}^{C}$ is a dimension-wise residual scaling factor
produced by the adaptive layer normalization (AdaLN) module~\cite{perez2018visual}. 
The \(\operatorname{AdaLN}\) layer encodes the input timestep \(t\) and the additional conditioning information \(c\) (e.g., class or text embedding) into channel-wise scale and shift parameters \(\gamma_k\) and \(\beta_k\). 
It then performs Adaptive Layer Normalization (AdaLN) on the hidden state \(z_t^k\):
\begin{equation}
\hat{z}_t^k = \bigl(1+\gamma_k\bigr)\,\operatorname{LayerNorm}(z_t^k) + \beta_k,
\label{AdaLN}
\end{equation}
where $\gamma_k, \beta_k$ are regressed by the MLP networks of AdaLN layer:
\begin{gather}
\gamma_k, \beta_k, \alpha_k = \operatorname{MLP}_k(t, c)
\label{regress}
\end{gather}
where $\alpha_k$ scales the $k$-th residual connection.
\subsection{Diffusion Transformer Representation}
Besides generating samples, a pretrained DiT can be used as a feature extractor by
forwarding a real image through the denoising Transformer and collecting intermediate
hidden states. Given an image $x$, we first obtain its latent $z_0=\mathcal{E}(x)$ and
add noise at a selected timestep $t$ to produce $z_t=z_0+\sigma(t)\epsilon$. We then
feed $z_t$ into the pretrained DiT under a fixed condition $c$ and extract the 
visual representation $F_{k,t}(x)$ from a chosen block $k$:
\begin{equation}
z_t^k=z_t^{k-1}+\alpha_{k-1} D_{k-1}(z_t^{k-1},t,c),\quad
F_{k,t}(x)=z_t^k .
\label{eq:dit_feature_extract}
\end{equation}
The resulting feature map $F_{k,t}(x)\in\mathbb{R}^{C\times H\times W}$ preserves the
spatial token layout of the latent representation and can be used as a dense visual
representation. The choice of timestep $t$, block index $k$, and feature normalization determines what visual information is
exposed by the representation.

\subsection{Diffusion Sampling and Classifier-Free Guidance}
Diffusion models~\cite{ho2020denoising,karras2022elucidating} generate samples by
progressively denoising an initial Gaussian variable $z_T \sim \mathcal{N}(0,I)$. Given
a clean sample $x \sim p_{\mathrm{data}}(x)$, the forward corruption process can be
written as $z_t=x+\sigma(t)\epsilon$, where $\sigma(t)$ is the noise schedule and
$\epsilon \sim \mathcal{N}(0,I)$. A denoising network $D_\theta(z_t,t,c)$ is trained to
predict the denoising direction conditioned on timestep $t$ and condition $c$, such as
a class label or text prompt. Classifier-Free Guidance (CFG)~\cite{ho2022classifier} improves conditional generation
by extrapolating from the unconditional prediction to the conditional prediction:
\begin{equation}
\hat{D}_{\theta}\left(z_t, c\right)=D_{\theta}\left(z_t,c\right)+w\left(D_{\theta}\left(z_t, c\right)-D_{\theta}\left(z_t,\hat{c}\right)\right)
\label{eq:cfg_prelim}
\end{equation}
where $\hat{c}$ denotes the unconditional input and $w$ is the CFG scale. The
difference $D_\theta(z_t,c)-D_\theta(z_t,\hat{c})$ captures the effect of the
external condition and therefore primarily strengthens prompt or class alignment.
However, semantic alignment and local detail fidelity are not identical objectives;
strong semantic guidance can still leave textures and small object parts
under-synthesized.

\section{Massive Activations in DiTs}
\label{sec:ma_analysis}
To systematically investigate the internal behavior of Diffusion 
Transformers (DiTs), we analyze the hidden states of several representative models, 
including SD3 and Flux. As shown in~\Cref{figure:f2}, 
we consistently observe the presence of \emph{Massive Activations} (MAs): 
a small subset of feature dimensions exhibits activation magnitudes that are 
substantially larger than those of the remaining dimensions across different models.
This phenomenon suggests that DiT representations are highly non-uniform 
and structured in magnitude distribution, raising the question of what 
functional role these massive activations play in model behavior. 
In the following sections, we conduct a detailed analysis to understand 
(i) where and how these activations emerge, and 
(ii) how they influence both visual generation and representation learning in DiTs.

\subsection{Spatial and Channel Properties of MAs}
\label{main:characteristics}
We first investigate where MAs appear in DiTs. As shown in~\Cref{figure:f3}, MAs appear
throughout the internal layers of DiTs, emerge early during training, and persist across
different model sizes. We observe the same layer-wise behavior in SD3, SD3.5, and Flux,
which further confirms that MAs are not isolated artifacts of a single architecture,
checkpoint, or layer.
We then examine their token and channel distributions. Unlike MAs 
in LLMs or ViTs~\cite{sun2024massive, darcet2023vision},
which often concentrate on special tokens or background/register-like tokens, MAs in
DiTs appear across all spatial patch tokens. At the same time, they are concentrated in
a few fixed feature dimensions, such as dimension 810 in SD3. This spatially
distributed but channel-concentrated pattern is consistent with the dense generative
objective of DiTs: every spatial token must participate in reconstructing local image
content, while a small set of feature dimensions can act as shared high-magnitude
channels for the internal computation of DiTs. This property makes MAs fundamentally
different from token-localized outliers in LLMs or ViTs and suggests that they may act
as shared activation components for dense visual processing.

\begin{figure}[t]
  \centering
  \includegraphics[width=\linewidth]{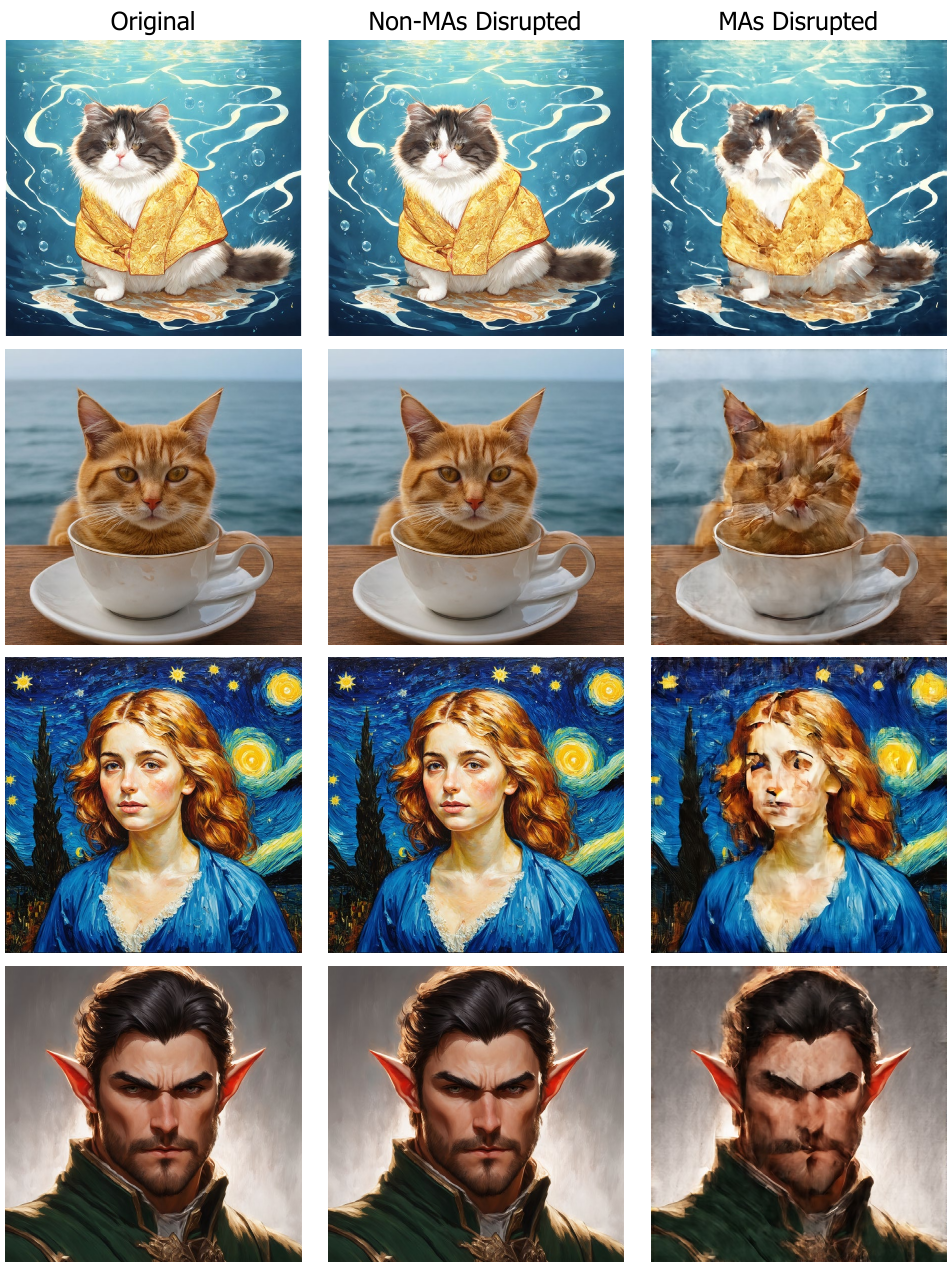}
  \caption{\textbf{Visual comparison of MA intervention.} 
  Disrupting massive activations (MAs) leads to a pronounced degradation in fine-grained image details, 
  whereas perturbing non-MA dimensions has negligible impact on the generated 
  outputs. This contrast indicates that MAs play a critical role in the synthesis 
  of visual details during the generative process.}
  \vspace{-3mm}
  \label{figure:ma_intervention}
\end{figure}

\begin{figure}[t]
  \centering
  \includegraphics[width =\linewidth]{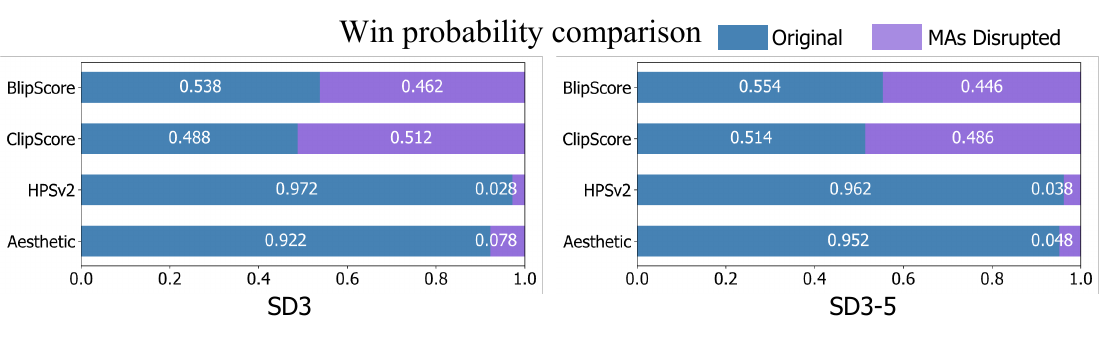}
  \caption{\textbf{Win probability comparison of MA intervention.} Win probability comparison for different models where we evaluate
  the model from two perspectives: \textbf{Prompt Alignment} (Blipscore and Clipscore)
  and \textbf{Local Detail Quality} (HPSv2.1 and Laion-Aesthetics). Disrupting massive
  activations markedly degrades the fidelity of fine-grained details in the generated
  images while exerting minimal impact on semantic content.}
  \vspace{-5mm}
  \label{figure:win_rate}
\end{figure}

\subsection{MAs Are Modulated by AdaLN}
\label{sec:ma_modulation}
To understand why MAs appear in fixed feature dimensions, we compare the hidden-state
activation distribution with the AdaLN residual scaling factor $\alpha_k$.
Recall from~\Cref{block} that the hidden state is updated as
\begin{equation}
z_t^{k+1}=z_t^k+\alpha_k D_k(z_t^k,t,c), \quad
\alpha_k=\operatorname{MLP}_k(t,c).
\label{computation}
\end{equation}
As shown in~\Cref{figure:f4}(a), peaks in $\alpha_k$ align with the dimensions where
hidden-state MAs concentrate. This indicates that the scaling factor $\alpha_k$ governs
the channel pattern of MAs. Since $\alpha_k$ is regressed from timestep $t$ and condition
embeddings $c$, we further analyze their respective impact.

As shown in~\Cref{figure:f4}(b), changing text prompts has little influence on the
magnitude of MAs; the activation values remain nearly constant across 1{,}000 prompts.
In contrast, the timestep has a strong effect: MA magnitudes progressively increase as
sampling moves from high-noise timesteps toward low-noise timesteps. Similar trends are
observed for SD3.5, Flux, and DiT-XL. These results suggest that MAs are primarily
stage-dependent rather than prompt-dependent. Since the timestep encodes the noise
level and denoising stage, MAs can be understood as activation components modulated by
the progression of the denoising trajectory rather than by the external semantic
condition.

\subsection{Generative Impact of MAs}

The structural and modulation properties discussed above reveal where 
MAs emerge and how they are controlled, but do not directly explain 
their functional role in DiTs. To investigate their impact on visual generation, 
we perform activation intervention~\cite{sun2024massive} by 
selectively disrupting MA dimensions in a single DiT layer and 
propagating the modified hidden states through the remaining blocks.

\paragraph{MAs are critical for visual detail synthesis.}
As shown in~\Cref{figure:ma_intervention}, disrupting MAs leads to a 
pronounced degradation of fine-grained local details, including textures 
and subtle object parts such as eyes and hair. In contrast, perturbing the 
same number of randomly selected non-MA dimensions has negligible impact 
on the generated outputs. This comparison indicates that MAs play a 
critical role in synthesizing local visual details during the generative 
process of DiTs.

\paragraph{MAs have limited influence on semantic content.}
Despite the degradation of local details, images generated with disrupted MAs 
largely preserve global semantics, including object identity, color composition, 
and overall layout, and remain semantically consistent with the original 
outputs~(\Cref{figure:ma_intervention}). This observation is further supported 
by prompt-alignment metrics: compared with the original outputs, the MA-disrupted 
model achieves comparable BLIPScore and CLIPScore win probabilities of 0.462 and 
0.512, respectively~(\Cref{figure:win_rate}). These results suggest that MAs 
primarily affect fine-grained visual detail rather than overall semantic content, 
which is consistent with our finding in~\Cref{sec:ma_modulation} that text 
embeddings have negligible influence on MAs.

\begin{figure}[t]
  \centering
  \includegraphics[width =\linewidth]{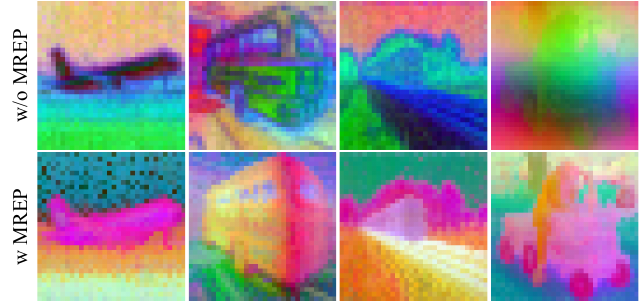}
  \vspace{-6mm}
  \caption{\textbf{Visual comparison with and without MREP.} 
  Raw DiT features w/o MREP exhibit poor semantic coherence and structural quality.
  Our MREP module effectively suppresses MA-dominated channels, 
  yielding more semantically coherent and spatially discriminative representations.}
    \vspace{-1mm}
  \label{figure:mrep}
\end{figure}

\subsection{Representational Impact of MAs}
\label{sec:ma_representation_impact}

The same activation structure leads to a distinct effect when DiTs 
are repurposed as feature extractors. Dense visual understanding requires 
spatial tokens to preserve discriminative local information, enabling features 
from different object parts or background regions to be reliably distinguished. 
However, since MAs concentrate in a few fixed dimensions and appear across nearly 
all spatial tokens, they introduce a shared high-magnitude direction into token 
representations. This dominant common direction makes features from different 
locations overly similar in the latent space, particularly under cosine-based 
comparison, while diminishing the contribution of locally informative non-MA 
dimensions. Consequently, raw DiT hidden states can become spatially 
indiscriminative and less effective for dense perception tasks, 
as shown in~\Cref{figure:mrep}.

\section{Methodology}
\label{sec:dam}
\begin{figure*}[htp]
  \centering
  \includegraphics[width =\linewidth]{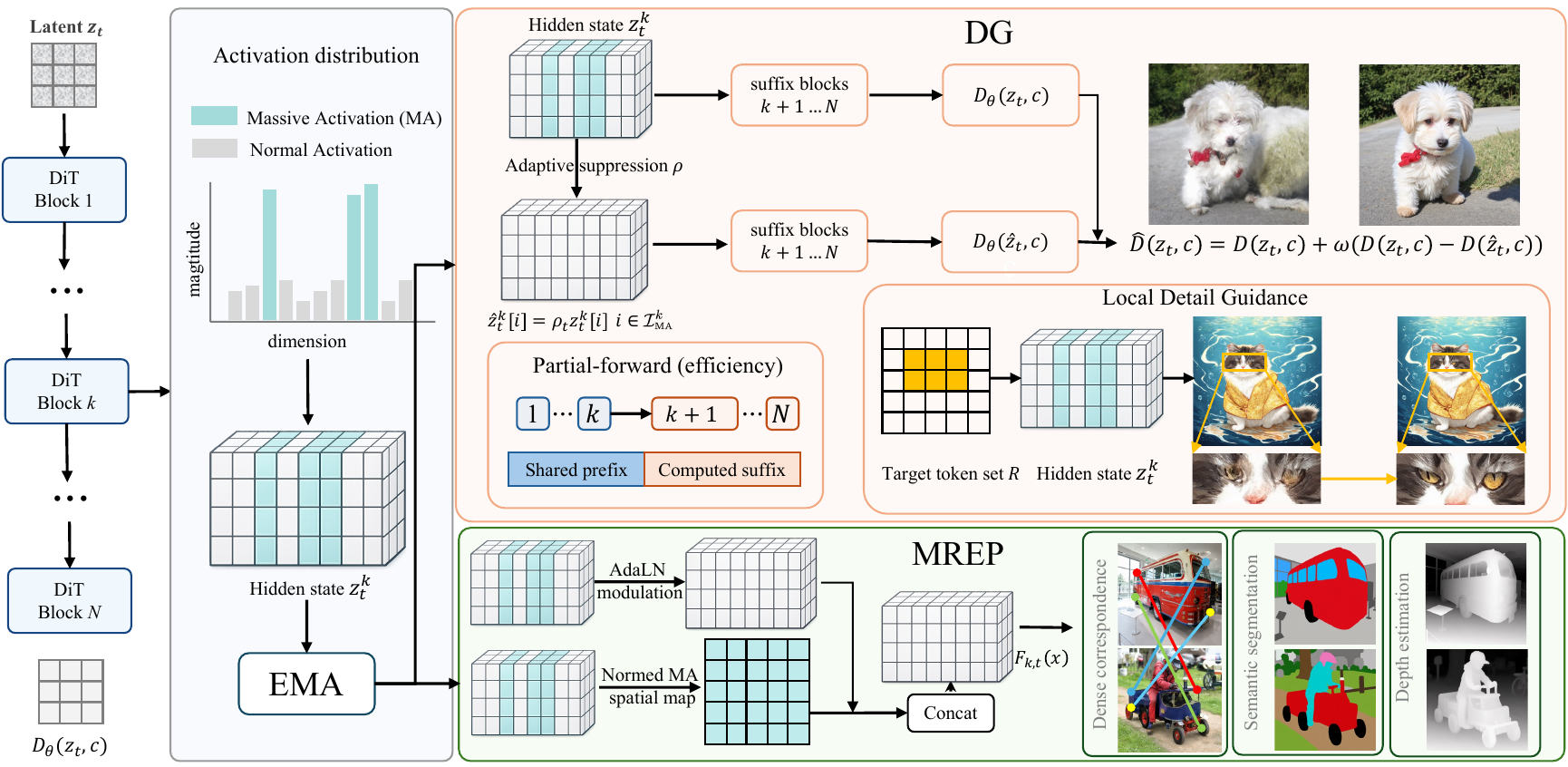}
  \vspace{-2mm}
\caption{\textbf{Overall framework of EMA.}
EMA leverages Massive Activations (MAs) as task-aware internal modulation signals 
in pretrained Diffusion Transformers (DiTs). For visual generation, 
EMA introduces Detail Guidance (DG), which constructs a detail-deficient 
counterfactual branch by suppressing MA dimensions and exploits the 
resulting residual signal to enhance fine-grained visual details. 
For visual understanding, EMA introduces MA-modulated Representation Extraction (MREP), 
which mitigates MA-dominated feature directions while preserving informative MA spatial structures, 
yielding more discriminative representations for dense perception tasks.}

  \vspace{-1mm}
  \label{figure:overall_framework}
\end{figure*}

\begin{table}[t]
\setlength\tabcolsep{1pt}
\color{black}
\centering
\caption{\textcolor{black}{\textbf{Unified formulations of different guidance methods.} 
$\hat{c}$ denotes the unconditional prompt, $\theta^*$ denotes under-capability
model, and $\hat{z}_t$ denotes $z_t$ with MAs-disrupted hidden state.}}
\label{tab:discussion}
\vspace{-2mm}
\resizebox{\linewidth}{!}{
\begin{tabular}{lcc}
\toprule Type & Formulation & Disruption \\
\midrule CFG&$\hat{D}_{\theta}\left(z_t, c\right)=D_{\theta}\left(z_t,c\right)+w\left(D_{\theta}\left(z_t, c\right)-D_{\theta}\left(z_t,\hat{c}\right)\right)$&$c$\\
 Autoguidance&$\hat{D}_{\theta}\left(z_t, c\right)=D_{\theta}\left(z_t,c\right)+w\left(D_{\theta}\left(z_t, c\right)-D_{{\theta}^*}\left(z_t,c\right)\right)$&$\theta$\\
 DG&$\hat{D}_{\theta}\left(z_t, c\right)=D_{\theta}\left(z_t,c\right)+w\left(D_{\theta}\left(z_t, c\right)-D_{\theta}\left(\hat{z}_t,c\right)\right)$&$z_t$\\
\bottomrule
\end{tabular}}
\end{table}
Based on the above analysis, MAs are not incidental numerical outliers, 
but structured and stage-dependent activation components that shape 
how DiTs process visual information. 
This section formally presents \textbf{EMA}, 
which leverages MAs to enhance both the generative and representational capabilities of DiTs, as shown in~\Cref{figure:overall_framework}. 
The core idea is to modulate MAs within the internal computations of DiTs, 
enabling targeted control over both the sampling process and representation extraction.
For generation, EMA introduces MA-driven \textbf{D}etail \textbf{G}uidance strategy (\textbf{DG}), 
which constructs a degraded, ``detail-deficient'' DiT by disrupting MAs and 
uses it to guide sampling toward detail-rich regions (see~\Cref{sec:detail_guidance,sec:partial_dg,sec:local_detail_guidance,sec:cfg_dg}). 
For understanding, EMA introduces \textbf{M}A-modulated \textbf{REP}resentation extraction method (\textbf{MREP}), 
which adaptively suppresses MA-dominated channels using the pretrained AdaLN operator, 
thereby recovering more spatially discriminative DiT representations (see~\Cref{sec:mrep_formulation}).

\subsection{Detail Guidance}
\label{sec:detail_guidance}
Motivated by the analysis in~\Cref{sec:ma_analysis}, we propose 
\textbf{D}etail \textbf{G}uidance (\textbf{DG}), a concise, 
training-free self-guidance strategy that leverages MAs to enhance 
fine-grained detail synthesis in DiTs. DG follows the general principle 
of self-guidance~\cite{karras2024guiding}, where sampling is guided by 
contrasting the original model with a deliberately degraded counterpart. 
Unlike generic degradation, DG requires a counterfactual branch that 
specifically weakens local details while preserving the semantic content 
of the original conditional prediction. This design ensures that the 
resulting residual provides a detail-specific guidance direction rather 
than an entangled semantic-detail signal.

To construct this branch, we disrupt MA patterns in the DiT hidden 
states of $z_t$. Formally, let $D_\theta$ denote the original pretrained 
DiT model, and let $z_t^k \in \mathbb{R}^{C \times H \times W}$ be the 
hidden state output of the \(k\)-th block. We obtain the modified hidden 
state $\hat{z}_t^k$ by adaptively suppressing MA dimensions in $z_t^k$ 
conditioned on timestep $t$, and then propagate it through the remaining 
blocks to produce the detail-deficient prediction.

\noindent\textbf{MA dimension selection.} For each candidate block $k$, we identify MA
dimensions from activation statistics collected once on a small calibration set. Let
$a_{k,i}$ denote the mean absolute activation magnitude of channel $i$ at block $k$,
averaged over calibration samples, timesteps, and spatial tokens. We define the median
channel magnitude at this block as
\begin{equation}
\mu_k=\operatorname{median}_{j}\;a_{k,j}.
\label{eq:ma_median}
\end{equation}
A channel is selected as an MA dimension if its average magnitude exceeds the median
magnitude by a ratio threshold $\tau_{\mathrm{MA}}$:
\begin{equation}
\mathcal{I}_{\mathrm{MA}}^k
=\left\{i\mid a_{k,i}>\tau_{\mathrm{MA}}\mu_k\right\}.
\label{eq:ma_selection}
\end{equation}
The selected MA dimensions are fixed after this calibration and reused for all prompts
during sampling. 

\noindent\textbf{Adaptive MA Suppression.} 
We suppress MAs adaptively along the denoising trajectory when constructing the
detail-deficient branch. Since fine visual details are also primarily formed in the late denoising stage (see~\Cref{figure:f4}),
the counterfactual branch should preserve MAs at early timesteps for global structure
formation, but strongly attenuate them at late timesteps to remove local detail
synthesis.
Let $k$ be the intervention block and $\mathcal{I}_{\mathrm{MA}}^k$ be its selected MA
dimensions. We define a timestep-dependent retention coefficient
$\rho_t\in[0,1]$ and apply it only to MA dimensions:
\begin{equation}
\hat{z}_{t,i}^{k}=
\begin{cases}
\rho_t z_{t,i}^{k}, & i\in\mathcal{I}_{\mathrm{MA}}^k,\\
z_{t,i}^{k}, & i\notin\mathcal{I}_{\mathrm{MA}}^k.
\end{cases}
\label{eq:adaptive_ma_suppression}
\end{equation}
Here $z_{t,i}^{k}$ denotes the activation values of channel $i$, with the coefficient
broadcast over spatial tokens. We use a sigmoid decay schedule over the normalized
timestep $t\in[0,1]$:
\begin{equation}
\rho_t=\sigma\!\left(\lambda_{\rho}(t-\tau_{\rho})\right),
\qquad
\sigma(x)=\frac{1}{1+\exp(-x)}.
\label{eq:rho_schedule}
\end{equation}
where $\tau_{\rho}=0.5$ controls the transition timestep and $\lambda_{\rho}=12$ controls
the sharpness of the decay.

\noindent\textbf{Detail Guidance formulation.}
After adaptive MA suppression, the modified hidden state $\hat{z}_t^k$ 
is propagated through the remaining DiT blocks, yielding a 
detail-deficient prediction $D_{\theta}(\hat{z}_t,c)$. 
The intervention block $k$ is selected via grid search. 
Since $D_{\theta}(\hat{z}_t,c)$ preserves the semantic content 
while weakening local details, the residual between the 
original and detail-deficient predictions provides a 
detail-specific guidance direction. DG is therefore defined as
\begin{equation}
\hat{D}_{\theta}(z_t,c)
=
D_{\theta}(z_t,c)
+
w\left(
D_{\theta}(z_t,c)-D_{\theta}(\hat{z}_t,c)
\right),
\label{eq:dg_from_branch}
\end{equation}
where $w$ controls the strength of detail guidance.

\subsection{Efficient Partial-Forward Detail Guidance}
\label{sec:partial_dg}
A naive implementation of DG would require two complete model evaluations: 
one for the original conditional prediction and another for the counterfactual 
detail-suppressed prediction. However, DG admits a more efficient 
partial-forward implementation. Since the counterfactual branch shares the 
same computation as the original branch before the intervention block $k$, 
we only need to recompute the intervened hidden state and the remaining DiT 
blocks after MA suppression. As a result, the additional inference cost is 
approximately proportional to $(N-k)/N$ of a single DiT forward pass, rather 
than a full extra evaluation as in standard guidance schemes such as CFG.
This design makes DG training-free, parameter-free, and computationally 
efficient. 

\subsection{Local Detail Guidance}
\label{sec:local_detail_guidance}
The spatially distributed nature of MAs further enables token-level control. While
DG suppresses selected MA dimensions over all spatial tokens, Local DG constructs the detail-deficient branch by suppressing MAs only
at target tokens. This produces a local counterfactual prediction in which fine
details are weakened inside the specified region, while the remaining tokens follow
the original DiT computation.

Formally, let $\mathcal{R}\subseteq\{1,\ldots,H\}\times\{1,\ldots,W\}$ denote the
target token set, which can be obtained from a mask, a bounding box, or selected
patches. For the intervention block $k$, we apply the same MA dimensions
$\mathcal{I}_{\mathrm{MA}}^k$ and retention coefficient $\rho_t$ as in DG, but only
for tokens in $\mathcal{R}$:
\begin{equation}
\hat{z}_{t,i,u}^{k}=
\begin{cases}
\rho_t z_{t,i,u}^{k}, & i\in\mathcal{I}_{\mathrm{MA}}^k,\; u\in\mathcal{R},\\
z_{t,i,u}^{k}, & \text{otherwise}.
\end{cases}
\label{eq:local_ma_suppression}
\end{equation}
Here $u$ indexes a spatial token. The modified hidden state is then propagated
through the remaining DiT blocks to obtain the local detail-deficient prediction
$D_{\theta}(\hat{z}_t^{\mathcal{R}},c)$. The resulting Local DG prediction is
\begin{equation}
\hat{D}_{\theta}^{\mathcal{R}}(z_t,c)
=
D_{\theta}(z_t,c)
+
w\left(
D_{\theta}(z_t,c)-D_{\theta}(\hat{z}_t^{\mathcal{R}},c)
\right),
\label{eq:local_dg}
\end{equation}
where the residual mainly reflects the missing details of the selected tokens.
Thus, Local DG keeps the training-free formulation of DG while turning MA
suppression into a token-specific detail-control mechanism. In practice, this
allows DG to serve as a local refinement operator: when a generated image contains
regions with insufficient texture or weak fine details, Local DG can selectively
enhance those regions while keeping the rest of the image largely unchanged.

\subsection{Integration with CFG.}
\label{sec:cfg_dg}
Classifier-free guidance (CFG)~\cite{ho2022classifier} is a standard technique that enhances semantic alignment 
by extrapolating between conditional and unconditional predictions. Our DG method is naturally complementary to CFG: 
whereas CFG strengthens semantic fidelity, DG explicitly enhances local detail quality (see~\Cref{tab:discussion}). 
The combined guidance is expressed as
\begin{equation}
\begin{split}
\hat{D}_{\theta}(z_t, c) &=
D_{\theta}(z_t, c) +
\lambda \big( D_{\theta}(z_t, c) - D_{\theta}(z_t, \hat{c}) \big) \\
&\quad + w \big( D_{\theta}(z_t, c) - D_{\theta}(\hat{z}_t, c) \big)
\end{split}
\end{equation}
where $\lambda$ and $w$ are the guidance scales of CFG and DG, respectively. 
$\hat{c}$ denotes the unconditional prompt and $\hat{z}_t$ denotes $z_t$ with MAs-disrupted hidden state

\begin{algorithm}[t]
\caption{Detail Guidance Sampling}
\label{alg:dg}
\begin{algorithmic}[1]
\REQUIRE pretrained DiT $D_\theta$, condition $c$, optional unconditional condition
$\hat{c}$, MA dimensions $\mathcal{I}_{\mathrm{MA}}^k$, intervention block $k$,
retention schedule $\rho_t$, DG scale $w$, optional CFG scale $\lambda$
\STATE Initialize noisy latent $z_T$.
\FOR{$t=T,\ldots,1$}
    \STATE $D_\theta(z_t,c),\,z_t^k\leftarrow
    \mathrm{Forward}(D_\theta,z_t,c;k)$
    \STATE $\hat{z}_{t,i}^{k}\leftarrow
    \begin{cases}
    \rho_t z_{t,i}^{k}, & i\in\mathcal{I}_{\mathrm{MA}}^k,\\
    z_{t,i}^{k}, & i\notin\mathcal{I}_{\mathrm{MA}}^k
    \end{cases}$
    \STATE $D_\theta(\hat{z}_t,c)\leftarrow
    D_\theta(z_t,c\,;\,z_t^k\leftarrow\hat{z}_t^k)$
    \STATE $g_{\mathrm{DG}}\leftarrow D_\theta(z_t,c)-D_\theta(\hat{z}_t,c)$
    \IF{CFG is used}
        \STATE $g_{\mathrm{CFG}}\leftarrow D_\theta(z_t,c)-D_\theta(z_t,\hat{c})$
        \STATE $\hat{D}_\theta(z_t,c)\leftarrow
        D_\theta(z_t,c)+\lambda g_{\mathrm{CFG}}+w g_{\mathrm{DG}}$
    \ELSE
        \STATE $\hat{D}_\theta(z_t,c)\leftarrow D_\theta(z_t,c)+w g_{\mathrm{DG}}$
    \ENDIF
    \STATE $z_{t-1}\leftarrow \mathrm{Sampler}(z_t,\hat{D}_\theta(z_t,c),t)$
\ENDFOR
\RETURN generated sample $z_0$.
\end{algorithmic}
\end{algorithm}

\subsection{MA-modulated Representation Extraction}
\label{sec:mrep_formulation}
Beyond generation, MAs also affect the quality of DiT representations. As shown
in~\Cref{sec:ma_representation_impact}, raw hidden states $z_t^k$ contain
MA-dominated channels shared by many spatial tokens, which weakens token-level
discrimination for dense understanding. Meanwhile, our analysis
in~\Cref{sec:ma_modulation} shows that these MA dimensions are closely aligned with
the AdaLN residual scaling factor $\alpha_k$. This suggests that the pretrained
AdaLN operator can be reused as a channel-wise modulator to suppress MA dominance in
features. Based on this observation, we introduce \textbf{M}A-modulated
\textbf{REP}resentation extraction (\textbf{MREP}), a training-free method that first
modulates DiT hidden states with AdaLN and then extracts dense visual representations.

\begin{figure}[t]
  \centering
  \includegraphics[width=\linewidth]{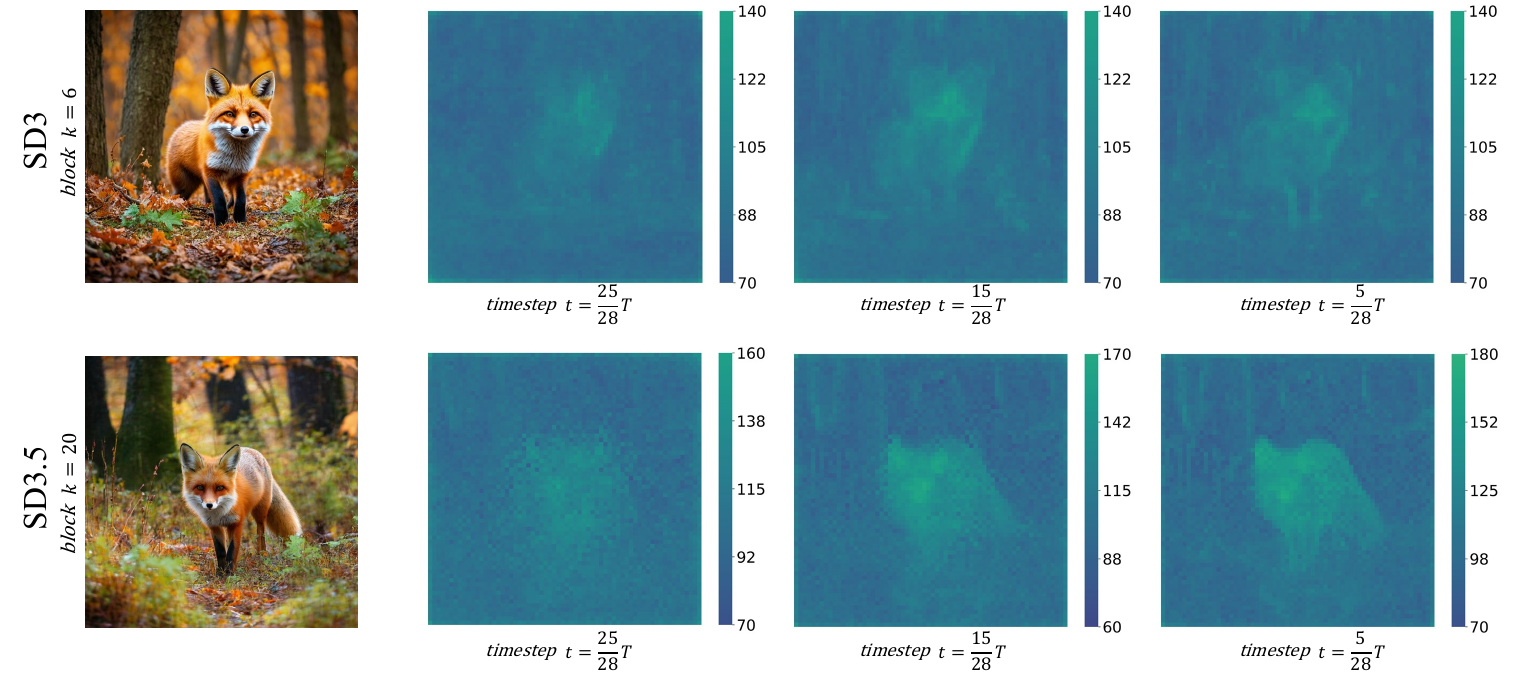}
\caption{\textbf{Visualization of MA spatial maps.}
MAs appear in fixed feature dimensions but produce 
structured responses across spatial tokens. 
This suggests that MA dimensions still encode useful spatial cues.}

  \vspace{-3mm}
  \label{figure:mrep_spatial}
\end{figure}

\noindent\textbf{Adaptive MA modulation with AdaLN.}
Given an image $x$, we encode it as $z_0=\mathcal{E}(x)$ and add noise at timestep
$t$ to obtain $z_t$. We then forward $z_t$ through the pretrained DiT and collect the
hidden state $z_t^k$ at block $k$. Instead of directly using $z_t^k$ as the visual
feature, MREP applies the pretrained AdaLN channel-wise weights from the same block.
Following~\Cref{AdaLN,regress}, the AdaLN parameters are
\begin{equation}
\gamma_k,\beta_k,\alpha_k=\operatorname{MLP}_k(t,c),
\label{eq:mrep_adaln_params}
\end{equation}
where $\gamma_k,\beta_k,\alpha_k\in\mathbb{R}^{C}$ are channel-wise parameters. Since
the MA dimensions are aligned with large responses in the AdaLN scaling factor
$\alpha_k$, these pretrained weights provide a built-in localization and modulation
signal for MA-dominated channels. We obtain the AdaLN-modulated hidden state as
\begin{equation}
\hat{z}_t^{k,\mathrm{Ada}}
=
\operatorname{AdaLN}(z_t^k;\gamma_k,\beta_k)
=
\left(1+\gamma_k\right)\operatorname{LayerNorm}(z_t^k)+\beta_k .
\label{eq:mrep_adaln_modulation}
\end{equation}
This channel-wise modulation reduces the dominance of MA channels while preserving
the spatial token structure needed for dense visual understanding.
After AdaLN modulation, MREP extracts representations from the modulated hidden 
state rather than the raw DiT hidden state:
\begin{equation}
\hat{F}_{k,t}^{\mathrm{Ada}}(x)=
\operatorname{Norm}\!\left(\hat{z}_t^{k,\mathrm{Ada}}\right),
\label{eq:mrep_feature}
\end{equation}
where $\operatorname{Norm}(\cdot)$ denotes the feature spatial normalization. 
This AdaLN-modulated branch suppresses the shared 
MA-dominated direction and better preserves token-level semantic differences, 
making DiT features more suitable for dense visual understanding.

\noindent\textbf{Preserving Spatial Structure in MA Dimensions.}
Although MA dimensions can dominate feature directions and weaken spatial 
discrimination, they are not purely detrimental to representation learning. 
As shown in~\Cref{figure:mrep_spatial}, MA channels still exhibit structured 
responses across visual tokens, indicating that they encode useful spatial cues. 
Therefore, instead of discarding MA dimensions entirely, MREP explicitly preserves 
their spatial structure as complementary information.

Specifically, we collect the spatial maps corresponding to the calibrated MA 
dimensions:
\begin{equation}
\mathcal{M}_{k,t}(x)
=
\left\{z_{t,i}^{k}\mid i\in\mathcal{I}_{\mathrm{MA}}^k\right\}.
\label{eq:mrep_ma_maps}
\end{equation}
However, due to their large magnitudes, directly using these MA maps can again 
dominate feature similarity. We therefore normalize them over spatial tokens:
\begin{equation}
\hat{\mathcal{M}}_{k,t}(x)
=
\operatorname{Norm}\!\left(\mathcal{M}_{k,t}(x)\right),
\label{eq:mrep_spatial_norm}
\end{equation}
where $\operatorname{Norm}(\cdot)$ denotes spatial normalization 
within each MA channel. This operation preserves the relative spatial response 
pattern of each MA channel while removing its excessive magnitude scale.

\begin{figure*}[t]
  \centering
  \includegraphics[width =\linewidth]{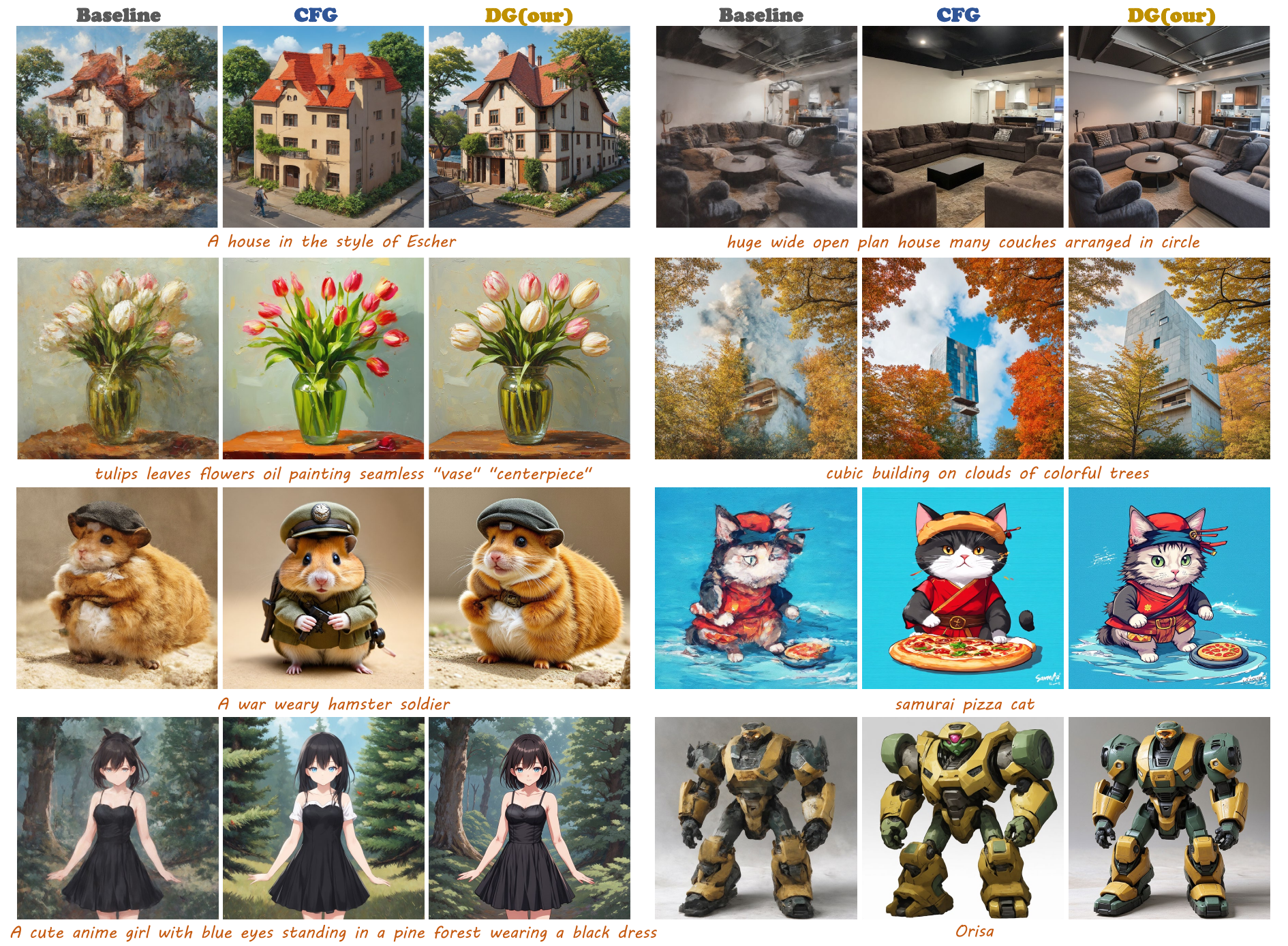}
  \caption{\textbf{Visual results of Detail Guidance (DG) on SD3.} Baseline denotes
  visual outputs without CFG. DG produces high-quality images with improved
  fine-grained details compared to Baseline. The CFG output is included as a reference
  for better comparison of detail quality.}
  \vspace{-5mm}
  \label{figure:f6}
\end{figure*}

\noindent\textbf{MREP Feature Integration.}
The final MREP representation combines the AdaLN-modulated representation with 
the normalized MA spatial maps:
\begin{equation}
\hat{F}_{k,t}(x)
=
\operatorname{Concat}\!\left(
\hat{F}_{k,t}^{\mathrm{Ada}}(x),\,
\hat{\mathcal{M}}_{k,t}(x)
\right).
\label{eq:mrep_final_feature}
\end{equation}
This design separates and reuses the two effects of MAs in a controlled manner. 
AdaLN modulation reduces the undesired directional dominance of MA dimensions, 
while spatial normalization preserves their useful spatial response patterns. 
As shown in~\Cref{figure:mrep}, by integrating these two complementary components, MREP produces DiT representations 
that are both less dominated by global MA directions and more informative for dense 
visual understanding.

\begin{table*}[t]
\setlength\tabcolsep{16pt}
\centering
\caption{\textbf{Evaluation of DG on dataset Pick- a-Pic.} We evaluate our
method on two generation settings: Conditional (Cond) and Classifier-Free guidance
(CFG). Our DG strategy brings substantial improvements on detail quality for both
settings, demonstrating its effectiveness in enhancing visual details. The best
highlights in bold.}
\label{tab:dg_pick}
\resizebox{\linewidth}{!}{
\begin{tabular}{ccccccc}
\toprule
\multirow{2}{*}{Model} & \multirow{2}{*}{Type} & \multirow{2}{*}{DG} & \multicolumn{2}{c}{Prompt Alignment} & \multicolumn{2}{c}{Detail Quality} \\
\cmidrule(lr){4-5}\cmidrule(lr){6-7}
& &  & Clipscore & Blipscore & HPSv2.1 & Aesthetic \\
\midrule
\multirow{4}{*}{SD3}& \multirow{2}{*}{Cond}& $\times$ & 22.11 & 66.74 & 21.84 &  5.58 \\
 && $\checkmark$ & \textbf{24.15} & \textbf{76.52} & \textbf{28.65} & \textbf{6.01}  \\
\cmidrule(lr){2-7}
&\multirow{2}{*}{CFG} & $\times$ & \textbf{26.64} & \textbf{87.01} & 28.23 & 5.80 \\
 && $\checkmark$ & 26.25 & 86.32 & \textbf{29.87} &\textbf{6.03} \\
\midrule
\multirow{4}{*}{SD3.5}& \multirow{2}{*}{Cond}& $\times$ & 24.90 & 70.09 & 23.65 &  5.94\\
 && $\checkmark$ & \textbf{26.01} & \textbf{83.66} & \textbf{29.23} & \textbf{6.16} \\
\cmidrule(lr){2-7}
&\multirow{2}{*}{CFG} & $\times$ & 27.67 & \textbf{92.62} & 29.9 & 6.01\\
 && $\checkmark$ & \textbf{27.68} & 91.61 & \textbf{30.7} & \textbf{6.18} \\
\midrule
\multirow{4}{*}{Flux1}& \multirow{2}{*}{Cond}& $\times$ & 22.09 & 57.60 & 19.33 &  5.50\\
 && $\checkmark$ & \textbf{25.69} & \textbf{80.55} & \textbf{27.88} & \textbf{6.13} \\
\cmidrule(lr){2-7}
&\multirow{2}{*}{CFG} & $\times$ & 27.04 & \textbf{87.76} & 29.16 & 5.96\\
 && $\checkmark$ & \textbf{27.14} & 86.23 & \textbf{29.25} & \textbf{6.12} \\
\midrule
\multirow{4}{*}{Flux2}& \multirow{2}{*}{Cond}& $\times$ & 24.84 & 75.36 & 26.92 & 6.24\\
 && $\checkmark$ & \textbf{25.78} & \textbf{81.47} & \textbf{29.18} & \textbf{6.39}\\
\cmidrule(lr){2-7}
&\multirow{2}{*}{CFG} & $\times$ & \textbf{28.18} & \textbf{91.86} & 30.42 & 6.33\\
 && $\checkmark$ & 28.12 & 91.04 & \textbf{31.06} & \textbf{6.45}\\
\bottomrule
\end{tabular}}
\vspace{-3mm}
\end{table*}

\begin{table*}[t]
\setlength\tabcolsep{4.5pt}
\centering
\caption{\textbf{Quantitative comparison with different sampling guidance.} We
evaluate HPSv2.1, T2I-CompBench, and Qalign on SD3. Higher is better for all
metrics. The best results within each model group are highlighted in bold.}
\label{tab:guidance}
\resizebox{\linewidth}{!}{
\begin{tabular}{lccccc ccc cc}
\toprule
\multirow{2}{*}{Method}
& \multicolumn{5}{c}{HPSv2.1 (\%) $\uparrow$}
& \multicolumn{3}{c}{T2I-CompBench (\%) $\uparrow$}
& \multicolumn{2}{c}{Qalign $\uparrow$} \\
\cmidrule(lr){2-6}\cmidrule(lr){7-9}\cmidrule(lr){10-11}
& Anime & Concept & Paint. & Photo & Avg.
& Color & Shape & Texture & HPSv2.1 & T2I-Comp. \\
\midrule
CFG & 31.55 & 30.87 & 31.22 & 28.27 & 30.48 & 53.61 & 51.20 & 52.45 & 4.66 & 4.74 \\
CFG++ & 31.57 & 30.76 & 30.96 & 27.54 & 30.21 & 46.39 & 47.18 & 46.33 & 4.68 & 4.73 \\
APG & 30.77 & 30.18 & 30.53 & 27.12 & 29.65 & 45.28 & 46.27 & 46.84 & 4.68 & 4.73 \\
CFG-Zero & 31.99 & 31.17 & 31.42 & 28.54 & 30.78 & 52.70 & 52.84 & 53.37 & 4.66 & \textbf{4.77} \\
$S^2$-Guidance & 32.14 & 31.32 & 31.70 & 29.19 & 31.09
& 59.63 & 58.71 & 56.77 & 4.65 & 4.74 \\
DG (Ours) & 32.14 & 31.05 & 31.50 & 29.09 & 30.95
& 58.62 & 58.06 & 57.97 & 4.66 & 4.75 \\
DG+CFG (Ours) & \textbf{32.94} & \textbf{31.78} & \textbf{31.71} & \textbf{29.85} & \textbf{31.57}
& \textbf{59.76} & \textbf{58.81} & \textbf{58.07} & \textbf{4.69} & 4.76 \\
\bottomrule
\end{tabular}}
\vspace{-7pt}
\end{table*}

\begin{table}[t]
\centering
\caption{\textbf{Performance comparison on ImageNet 256x256 with DiT-XL/2.} Prec. and
Rec. denote precision and recall.}
\label{tab:imagenet}
\resizebox{\linewidth}{!}{
\begin{tabular}{cccccc}
\toprule
Type & DG & FID $\downarrow$ & IS $\uparrow$ & Prec. $\uparrow$ & Rec. $\uparrow$ \\
\midrule
\multirow{2}{*}{Uncond}& $\times$ & 16.95 & 105.64 & 0.61 & 0.76 \\
& $\checkmark$ & 9.68 & 122.22 & 0.66 & 0.67 \\
\midrule
\multirow{2}{*}{Cond} & $\times$ & 9.52 & 122.79 & 0.66 & 0.63 \\
& $\checkmark$ & 5.77 & 179.26 & 0.78 & 0.55 \\
\bottomrule
\end{tabular}}
\vspace{-3mm}
\end{table}

\section{Experiments}
\label{sec:experiments}
\subsection{Experimental Setup}

\noindent\textbf{Model variants.}
We evaluate EMA across a broad range of image and video Diffusion Transformers. 
For text-to-image generation, we consider four representative large-scale 
DiT-based models: SD3-Medium~\cite{esser2024scaling} (SD3), SD3.5-Large~\cite{esser2024scaling} (SD3.5), 
Flux1~\cite{Flux}, and Flux2~\cite{Flux}. Since Flux is released as a CFG-distilled model, we use its 
de-distilled variant when evaluating DG independently from CFG. To examine 
whether DG depends on a particular DiT implementation, we further test it on 
PixArt-$\alpha$~\cite{chen2023pixart}, which uses shared AdaLN layers, and 
SANA~\cite{xie2024sana}, which adopts linear attention and convolutional 
local modeling. For text-to-video generation, we evaluate EMA on two video 
DiTs: Wan-1.3B, and Wan-14B~\cite{wan2025wan}. 
In addition, we assess EMA on class-conditional generation with 
DiT-XL/2~\cite{peebles2023scalable} on ImageNet at $256\times256$ resolution.

\noindent\textbf{Visual generation evaluation.}
For text-to-image generation, we evaluate EMA on two standard prompt benchmarks: 
the Pick-a-Pic ``test unique'' split~\cite{kirstain2023pick} and HPSv2.1~\cite{wu2023human}. 
To quantify prompt alignment, we compute CLIPScore~\cite{radford2021learning} 
using \textit{clip-vit-large-patch14} and BLIPScore~\cite{li2022blip} 
using \textit{blip-itm-large-coco}. CLIPScore measures global text-image 
semantic consistency, while BLIPScore provides a complementary image-text 
matching signal. To assess perceptual quality and visual preference, we 
report HPSv2.1~\cite{wu2023human} and LAION-Aesthetics~\cite{schuhmann2022laionaesthetics}. 
For ImageNet class-conditional generation, we follow the standard DiT evaluation protocol 
and report FID, IS, Precision, and Recall. For text-to-video generation, 
we adopt the standard prompts and evaluation metrics from VBench~\cite{huang2024vbench}.

\noindent\textbf{Visual understanding evaluation.}
To assess the representation quality of EMA, we conduct experiments on three dense 
visual perception tasks: semantic correspondence, semantic segmentation, and depth 
estimation. For semantic correspondence, we evaluate on three widely used benchmarks: 
SPair-71k~\cite{min2019spair}, PF-Pascal~\cite{ham2016proposal}, and 
AP-10K~\cite{zhang2024telling}. AP-10K is a large-scale semantic correspondence 
benchmark built upon the AP-10K dataset~\cite{yu2021ap}, comprising 2.61 million 
training pairs and 36,000 test pairs across three settings: intra-species, 
cross-species, and cross-family matching. Following prior work~\cite{tang2023emergent, zhang2024tale}, 
we report the percentage of correct keypoints (PCK). 
For semantic segmentation, we evaluate on ADE20k~\cite{zhou2017scene} and report mean 
intersection over union (mIoU). For depth estimation, we evaluate on 
NYUv2~\cite{silberman2012indoor} and report standard metrics, including 
absolute relative error (Abs Rel), root mean squared error (RMSE), and threshold accuracy $\delta_i$.

\noindent\textbf{Computational resources.}
All quantitative experiments are conducted on a single NVIDIA L40S GPU with 48GB memory. 
To measure computational overhead, we generate 100 
images at $1024\times1024$ resolution and report the average latency and memory usage.

\begin{figure*}[tp]
  \centering
  \includegraphics[width =\linewidth]{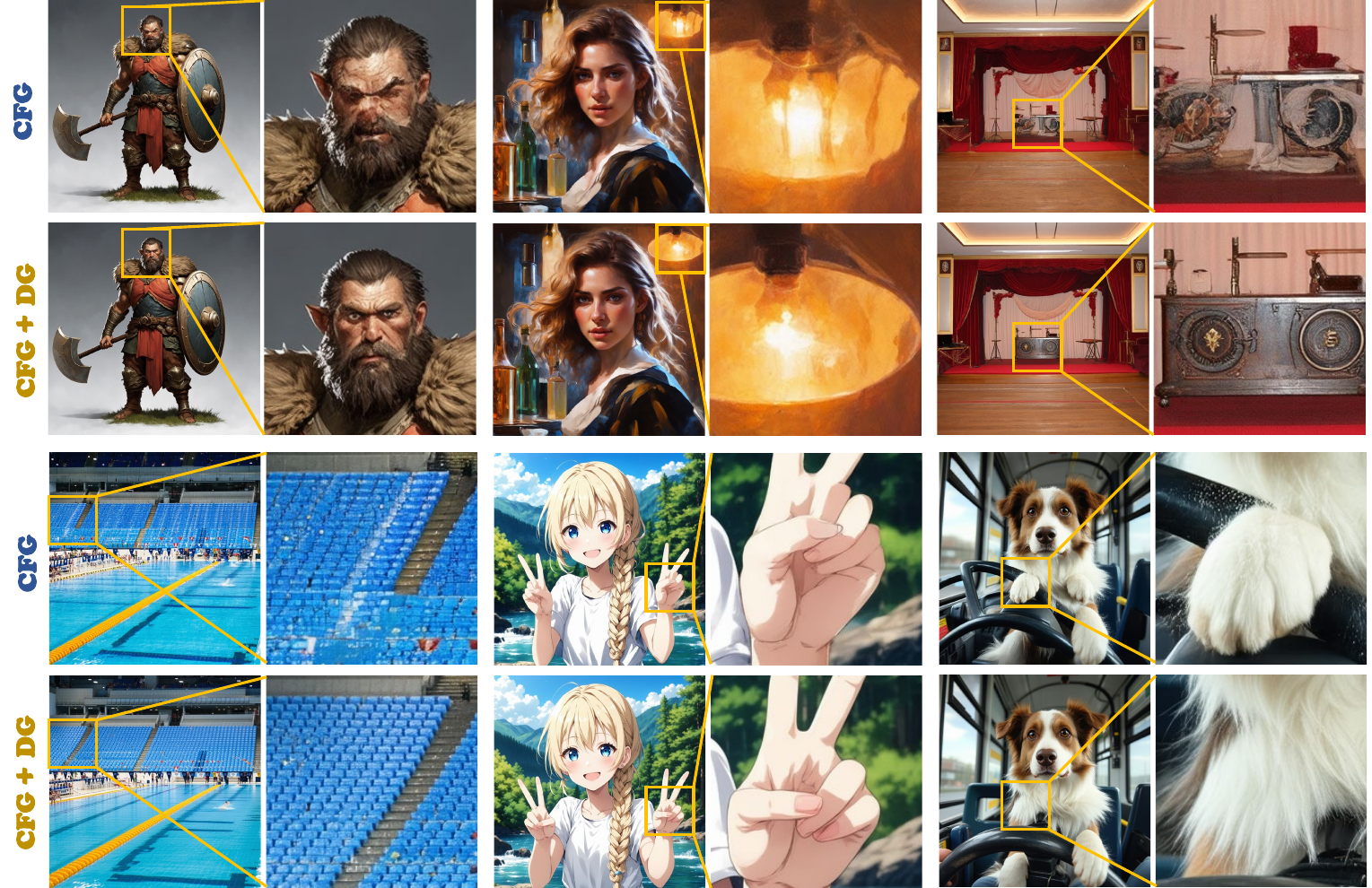}
  \vspace{-5mm}
  \caption{\textbf{Integration with CFG.} Rows 1-2: SD3; Rows 3-4: SD3.5. Incorporating
  DG into Classifier-Free Guidance (CFG) yields outputs with markedly richer and more
  refined local details.}
  \vspace{-3mm}
  \label{figure:f7}
\end{figure*}

\vspace{-2mm}
\subsection{Main Results on Image Generation}
\vspace{-1mm}
\subsubsection{Text-to-image generation}

\noindent\textbf{Evaluation of DG.}
\Cref{tab:dg_pick} summarizes the performance of Detail Guidance (DG) on three pretrained DiTs. 
DG consistently improves both prompt alignment and fine-grained detail quality across all models. 
For example, on SD3.5, DG increases the BlipScore from 70.09 to 83.66 while improving the 
Aesthetic score from 5.94 to 6.16. The qualitative comparisons in~\Cref{figure:f6} further 
demonstrate that DG enhances local textures and structural details while preserving the overall 
image composition. Notably, the consistent improvements on Flux indicate that DG generalizes 
beyond the SD3 family and remains effective for stronger large-scale DiTs.
Compared with CFG, DG achieves higher detail-quality metrics (e.g., Aesthetic 6.01 vs.\ 5.80 on SD3), 
whereas CFG provides stronger prompt alignment. As illustrated in~\Cref{figure:f6}, 
DG produces richer local textures, while CFG better preserves semantic consistency with the 
text prompt. These observations suggest that the two guidance strategies improve complementary 
aspects of generation: DG primarily enhances fine-grained detail synthesis, whereas 
CFG strengthens semantic alignment.

\noindent\textbf{Integrating DG with CFG.}
DG is fully compatible with CFG and consistently improves detail quality when combined with it, 
as shown in~\Cref{tab:dg_pick}. The visual results in~\Cref{figure:f7} further show that the 
combined guidance produces more refined local structures and higher overall image quality. 
This complementary effect is also reflected quantitatively: CFG mainly improves 
prompt alignment, whereas DG consistently boosts detail-quality metrics.

\noindent\textbf{Comparison with state-of-the-art.}
We further compare DG with representative guidance methods, 
including APG~\cite{sadat2024eliminating}, CFG++~\cite{chung2024cfg++}, 
CFG-Zero~\cite{fan2025cfg}, and $S^2$-Guidance~\cite{chen2025s}. 
As reported in~\Cref{tab:guidance}, DG achieves the highest Aesthetic score among 
standalone guidance methods without requiring an unconditional branch. Compared with PAG, 
which constructs a degraded branch by perturbing attention maps, 
DG delivers superior detail-quality performance. Moreover, when combined with CFG, 
DG achieves the best average HPSv2.1 score, further validating our 
central hypothesis that DG provides a detail-oriented guidance direction 
that complements semantic guidance from CFG.

\begin{table*}[t]
\setlength\tabcolsep{4pt}
\centering
\caption{\textbf{Evaluation on video DiTs.} We compare DG with representative guidance
methods on Wan1.3B and Wan14B. Higher is better for all metrics. The best results within
each model group are highlighted in bold, while the second best is underlined.}
\label{tab:video_dit}
\resizebox{\linewidth}{!}{
\begin{tabular}{ccccccccccc}
\toprule
\multirow{2}{*}{Model} & \multirow{2}{*}{Method}
& Total & Quality & Semantic & Subject & Background & Aesthetic & Imaging & Object & Appearance \\
& & Score & Score & Score & Consistency & Consistency & Quality & Quality & Class & Style \\
\midrule
\multirow{8}{*}{Wan1.3B}
& CFG & 80.29 & 84.32 & 64.16 & 96.53 & 95.46 & 60.52 & 67.65 & 77.06 & 20.15 \\
& CFG++ & 80.35 & 83.58 & \textbf{67.43} & \textbf{96.70} & 93.28 & 59.02 & \textbf{69.14} & 70.06 & 19.75 \\
& APG & 70.83 & 77.13 & 45.61 & 96.45 & 95.39 & 49.42 & 64.39 & 59.02 & 20.01 \\
& STG & 78.78 & 83.92 & 58.19 & 95.03 & 96.04 & 59.03 & 65.59 & 68.20 & 21.51 \\
& CFG-Zero & 80.71 & 84.51 & 65.53 & 96.33 & 94.56 & 59.69 & 69.05 & 78.16 & 20.31 \\
& $S^2$-Guidance & 80.93 & 84.74 & 65.70 & 96.57 & 
95.80 & 60.52 & 68.19 & 78.09 & 20.59 \\
& DG (Ours) & \underline{81.02} & \underline{84.86} & \underline{66.12} & 96.61 & 
\underline{95.96} & \underline{60.61} & 68.96 & \underline{78.33} & \underline{21.06} \\
& DG+CFG (Ours) & \textbf{81.16} & \textbf{84.97} & 66.35 & \underline{96.66} & 
\textbf{96.12} & \textbf{60.74} & \underline{69.12} & \textbf{78.51} & \textbf{21.24} \\
\midrule
\multirow{4}{*}{Wan14B}
& CFG & 82.65 & 84.88 & 73.76 & \textbf{94.45} & \textbf{97.66} & 68.68 & \textbf{67.82} & 84.97 & 22.14 \\
& $S^2$-Guidance & 82.84 & 84.89 & 74.65 & 94.21 & 97.56 & 68.78 & 67.77 & 89.08 & 22.27 \\
& DG (Ours) & \underline{82.93} & \underline{84.96} & \underline{74.82} & 94.30 & 97.59 & \underline{68.86} & 67.79 & \underline{89.31} & \underline{22.34} \\
& DG+CFG (Ours) & \textbf{83.05} & \textbf{85.04} & \textbf{75.06} & \underline{94.36} & \underline{97.62} & \textbf{68.94} & \underline{67.81} & \textbf{89.52} & \textbf{22.46} \\
\bottomrule
\end{tabular}}
\vspace{-5pt}
\end{table*}

\subsubsection{Class-to-image generation}
We further evaluate whether DG transfers from text-conditioned generation to
class-to-image generation. As reported in~\Cref{tab:imagenet}, 
DG consistently improves the main sample
quality metrics under both unconditional and class-conditional sampling. In the
unconditional setting, DG reduces FID from 16.95 to 9.68. Under
class conditioning, the improvement becomes more pronounced: FID decreases from
9.52 to 5.77 and IS increases from 122.79 to 179.26. 
These results indicate that DG is not tied to text prompts or
text-image alignment metrics; instead, suppressing detail-deficient MA
directions provides a broadly useful refinement signal for DiT-based visual
generation.

\subsection{Main Results on Video Generation}

We further evaluate DG on Wan video models~\cite{wan2025}. 
Compared with image generation, video generation poses a more challenging setting, 
as enhancing local details must also preserve temporal consistency across frames.
As reported in~\Cref{tab:video_dit}, DG consistently improves performance across 
both model scales. On Wan1.3B, DG outperforms standard CFG and other competing 
guidance methods in terms of the overall score, while DG+CFG achieves the best 
overall performance. A similar trend is observed on Wan14B, where DG surpasses the 
strong Wan14B baseline, and DG+CFG attains the highest total, quality, semantic, 
aesthetic, object, and appearance scores. These consistent improvements demonstrate 
that the MA-based guidance is effective not only for spatial tokens in image DiTs 
but also for spatiotemporal tokens in video DiTs.
The comparison between DG and DG+CFG further reveals their complementary roles. CFG 
primarily improves prompt adherence and high-level semantic alignment, whereas 
DG enhances fine-grained visual details and appearance fidelity.

\subsection{Main Results on Visual Understanding}
\subsubsection{Semantic correspondence}
We first evaluate whether MREP improves the spatial discriminability of DiT
features on semantic correspondence. This task is particularly sensitive to
whether the representation preserves object-part geometry, since the model must
match keypoints across instances, categories, and appearance changes. As shown
in~\Cref{tab:semantic_correspondence}, the raw DiTF representation already forms
a strong unsupervised baseline, especially on PF-Pascal. However, removing the
MREP modulation leads to a substantial degradation across all datasets, showing
that MA-dominated directions can obscure spatially meaningful correspondences.
With MREP, the model consistently improves over DiTF under the unsupervised
setting, achieving the best results on SPair-71k, AP-10K, and PF-Pascal. These
results suggest that MREP does not merely improve global image semantics; it
also strengthens the local part-level structure required for dense matching.

\begin{table*}[t]
\setlength\tabcolsep{3.5pt}
\centering
\caption{\textbf{Semantic correspondence on different datasets.} We
report PCK on SPair-71k, AP-10K intra-species (I.S.), AP-10K cross-species (C.S.),
AP-10K cross-family (C.F.), and PF-Pascal. S and U denote supervised and unsupervised
settings. The best results are in
bold, and the second-best results are underlined.}
\label{tab:semantic_correspondence}
\resizebox{\linewidth}{!}{
\begin{tabular}{clccccccccccccccc}
\toprule
& \multirow{2}{*}{Method}
& \multicolumn{3}{c}{SPair-71k}
& \multicolumn{3}{c}{AP-10K-I.S.}
& \multicolumn{3}{c}{AP-10K-C.S.}
& \multicolumn{3}{c}{AP-10K-C.F.}
& \multicolumn{3}{c}{PF-Pascal}\\
\cmidrule(lr){3-5}\cmidrule(lr){6-8}\cmidrule(lr){9-11}\cmidrule(lr){12-14}\cmidrule(lr){15-17}
& & 0.01 & 0.05 & 0.10 & 0.01 & 0.05 & 0.10 & 0.01 & 0.05 & 0.10
& 0.01 & 0.05 & 0.10 & 0.05 & 0.10 & 0.15\\
\midrule
\multirow{4}{*}{S}
& SCorrSAN~\cite{huang2022learning} & 3.6 & 36.3 & 55.3 & -- & -- & -- & -- & -- & -- & -- & -- & -- & 81.5 & 93.3 & 96.6\\
& CATs++~\cite{cho2022cats++} & 4.3 & 40.7 & 59.8 & -- & -- & -- & -- & -- & -- & -- & -- & -- & 84.9 & 93.8 & 96.8\\
& DHF~\cite{luo2024diffusion} & 8.7 & 50.2 & 64.9 & 8.0 & 45.8 & 62.7 & 6.8 & 42.4 & 60.0 & 5.0 & 32.7 & 47.8 & 78.0 & 90.4 & 94.1\\
& SD+DINO (S)~\cite{zhang2024tale} & 9.6 & 57.7 & 74.6 & 9.9 & 57.0 & 77.0 & 8.8 & 53.9 & 74.0 & 6.9 & 46.2 & 65.8 & 80.9 & 93.6 & 96.9\\
\midrule
\multirow{5}{*}{U}
& DINOv2+NN~\cite{oquab2023dinov2} & 6.3 & 38.4 & 53.9 & 6.4 & 41.0 & 60.9 & 5.3 & 37.0 & 57.3 & 4.4 & 29.4 & 47.4 & 63.0 & 79.2 & 85.1\\
& DIFT~\cite{tang2023emergent} & 7.2 & 39.7 & 52.9 & 6.2 & 34.8 & 50.3 & 5.1 & 30.8 & 46.0 & 3.7 & 22.4 & 35.0 & 66.0 & 81.1 & 87.2\\
& DiTF~\cite{gan2025unleashing} & \underline{8.3} & \underline{49.3} & \underline{64.0} & \underline{7.6} & \underline{49.7} & \underline{62.2} & \underline{6.8} & \underline{48.2} & \underline{61.7} & \underline{5.7} & \underline{34.2} & \underline{48.9} & \underline{89.5} & \underline{95.8} & \underline{97.6}\\
\cmidrule(lr){2-17}
& w/o MREP & 3.5 & 21.4 & 29.9 & 3.5 & 22.4 & 32.2 & 2.9 & 19.9 & 29.8 & 2.3 & 14.2 & 22.1 & 43.4 & 52.0 & 54.8\\
& MREP (Ours) & \textbf{8.6} & \textbf{50.1} & \textbf{64.7} & \textbf{7.8} & \textbf{50.4} & \textbf{63.0} & \textbf{7.0} & \textbf{49.0} & \textbf{62.5} & \textbf{5.9} & \textbf{35.0} & \textbf{49.6} & \textbf{90.1} & \textbf{96.1} & \textbf{97.8}\\
\bottomrule
\end{tabular}}
\vspace{-3mm}
\end{table*}

\subsubsection{Semantic segmentation}
We further evaluate the learned representations on ADE20K semantic segmentation.
Compared with semantic correspondence, segmentation requires the representation
to support dense category prediction over the entire image, making it a useful
test of whether MREP preserves both local boundaries and semantic region
consistency. As reported in~\Cref{tab:segmentation_pretraining}, the MREP-enhanced
features achieve 54.9 single-scale mIoU and 56.1 multi-scale mIoU, outperforming
the raw DiTF features and remaining competitive with strong diffusion-based
pre-training baselines. In contrast, the variant without MREP drops notably,
which again indicates that directly using MA-dominated DiT features can hurt
dense perception. These gains show that MREP provides a task-useful
representation refinement for semantic understanding, rather than only improving
feature matching on correspondence benchmarks.

\begin{table}[t]
\centering
\caption{\textbf{Semantic segmentation on dataset ADE20k.} We report
mIoU under single-scale (SS) and multi-scale (MS) testing.}
\label{tab:segmentation_pretraining}
\normalsize
\renewcommand{\arraystretch}{1.08}
\begin{tabular*}{\linewidth}{@{}l@{\extracolsep{\fill}}cc@{}}
\toprule
Method & mIoU$^{\mathrm{SS}}$ & mIoU$^{\mathrm{MS}}$ \\
\midrule
\multicolumn{3}{@{}l}{\textit{self-supervised pre-training}}\\
DINOv2~\cite{oquab2023dinov2} & 47.7 & 53.1\\
\midrule
\multicolumn{3}{@{}l}{\textit{SD-based pre-training}}\\
VPD~\cite{zhao2023unleashing} & 53.7 & 54.6\\
VPD(R) & 53.1 & 54.2\\
VPD(LS) & 53.7 & 54.4\\
TADP & 54.8 & 55.9\\
\midrule
\multicolumn{3}{@{}l}{\textit{DiTs-based pre-training}}\\
$\mathrm{DiTF}$ & 53.6 & 54.8\\
\midrule
w/o MREP & 48.8 & 51.2\\
MREP (Ours) & \textbf{54.9} & \textbf{56.1}\\
\bottomrule
\end{tabular*}
\vspace{-2mm}
\end{table}

\subsubsection{Depth estimation}
Finally, we evaluate MREP on NYUv2 depth estimation to test whether the same
representation benefits geometric prediction. Depth estimation differs from the
previous semantic tasks because it relies more strongly on layout, object scale,
and 3D structure. As shown in~\Cref{tab:depth_estimation}, MREP obtains the best
overall performance among the compared methods, reducing RMSE to 0.220 and REL
to 0.060 while improving threshold accuracy. The comparison with the w/o MREP
variant is especially informative: without the modulation, performance falls
behind strong diffusion-based baselines, whereas MREP yields clear gains across
both error and accuracy metrics. This consistency across correspondence,
segmentation, and depth estimation supports our hypothesis that modulating
massive activations helps convert generative DiT features into more reliable
representations for visual understanding.

\begin{table}[t]
\centering
\caption{\textbf{Depth estimation on dataset NYUv2.} We report RMSE,
threshold accuracy $\delta_i$, and relative error (REL).}
\label{tab:depth_estimation}
\normalsize
\renewcommand{\arraystretch}{1.08}
\setlength\tabcolsep{2pt}
\begin{tabular*}{\linewidth}{@{}l@{\extracolsep{\fill}}ccccc@{}}
\toprule
Method & RMSE$\downarrow$ & $\delta_1\uparrow$ & $\delta_2\uparrow$ & $\delta_3\uparrow$ & REL$\downarrow$ \\
\midrule
SwinV2-L & 0.287 & 0.949 & 0.994 & 0.999 & 0.083\\
AiT & 0.275 & 0.954 & 0.994 & 0.999 & 0.076\\
ZoeDepth & 0.270 & 0.955 & 0.995 & 0.999 & 0.075\\
VPD~\cite{zhao2023unleashing} & 0.254 & 0.964 & 0.995 & 0.999 & 0.069\\
VPD(R) & 0.248 & 0.965 & 0.995 & 0.999 & 0.068\\
VPD(LS) & 0.235 & 0.971 & 0.996 & 0.999 & 0.064\\
TADP & 0.225 & 0.976 & 0.997 & 0.999 & 0.062\\
\midrule
w/o MREP & 0.256 & 0.951 & 0.994 & 0.999 & 0.071\\
MREP (Ours) & \textbf{0.220} & \textbf{0.978} & \textbf{0.997} & \textbf{0.999} & \textbf{0.060}\\
\bottomrule
\end{tabular*}
\vspace{-2mm}
\end{table}

\subsection{Additional Applications}
\subsubsection{Efficient Partial-Forward Detail Guidance.}

We evaluate the computational efficiency of DG using the partial-forward implementation 
described in~\Cref{sec:partial_dg}. As reported in~\Cref{tab:overhead}, 
DG consistently reduces inference latency relative to CFG 
while achieving higher aesthetic quality. For example, on SD3.5, 
DG reduces the inference time from 15.7\,s to 10.6\,s while 
improving the Aesthetic score from 6.01 to 6.16. 
These results demonstrate that DG offers a favorable 
efficiency-quality trade-off, enhancing visual fidelity without 
incurring the full computational overhead of an additional denoising branch.

\subsubsection{DG on Non-DiT Architectures.}

To examine whether DG relies on the standard DiT architecture, 
we further evaluate it on PixArt-alpha~\cite{chen2023pixart} 
and SANA~\cite{xie2024sana}. PixArt-alpha injects text conditioning 
via cross-attention with shared AdaLN layers, whereas SANA replaces 
standard attention with linear attention and incorporates local 
convolutional operations. Following the same MA suppression strategy, 
we suppress dimension in PixArt-alpha and SANA. 
As shown in~\Cref{tab:non_dit}, DG consistently improves detail 
quality under both conditional generation and CFG. 
These results indicate that DG captures a general detail-guidance 
mechanism rather than exploiting architectural characteristics specific to DiTs.

\subsubsection{Token-specific Local DG.}

\begin{figure}[t]
  \centering
  \includegraphics[width=0.95\linewidth]{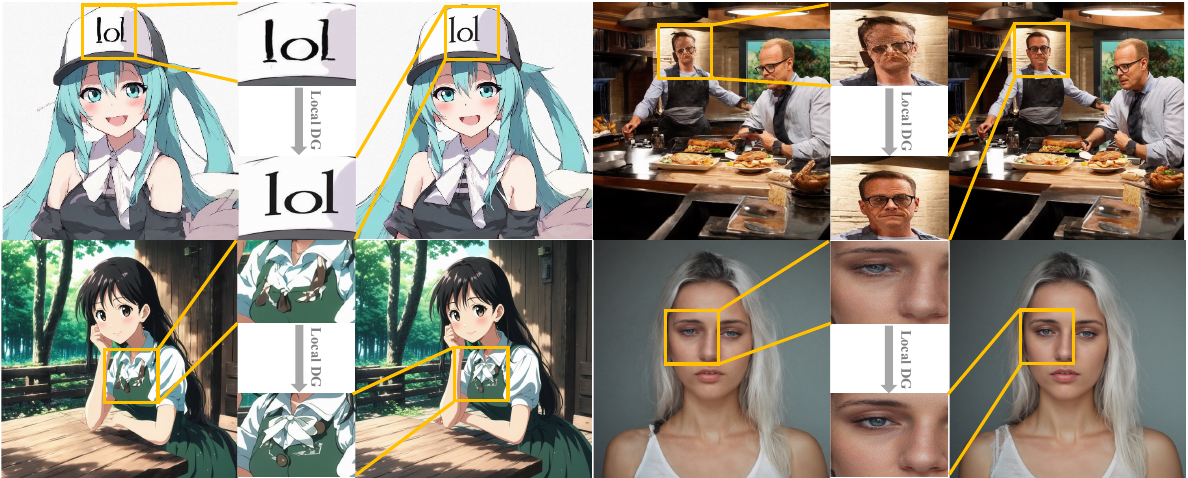}
  \caption{\textbf{Visual results of Local DG.} Selectively applying DG to MAs in target spatial
  tokens enhances the details of the selected region while keeping other regions
  largely unchanged.}
  \vspace{-3mm}
  \label{figure:dg_local_main}
\end{figure}

Our spatial analysis of MAs suggests that they function as token-wise modulators of 
local detail synthesis. To validate this hypothesis, we construct Local 
DG by suppressing MAs only for spatial tokens corresponding to a target region, 
instead of applying suppression to all spatial tokens. 
As illustrated in~\Cref{figure:dg_local_main}, Local DG selectively enhances 
details within the target region while leaving surrounding objects and the 
background largely unchanged. This result further supports the interpretation 
that DG provides spatially controllable detail guidance.

\begin{table}[t]
\centering
\caption{\textbf{Computational overhead of efficient partial-forward DG.} Latency is measured for 1024x1024
generation on one NVIDIA L40S GPU.}
\label{tab:overhead}
\resizebox{\linewidth}{!}{
\begin{tabular}{ccccc}
\toprule
Model & Type & Memory (GB) & Latency (s) & Aesthetic\\
\midrule
\multirow{3}{*}{SD3} & Cond & 17 & 2.3 & 5.58\\
& CFG & 20 & 4.3 & 5.80\\
& DG & 20 & 3.5 & 6.01\\
\midrule
\multirow{3}{*}{SD3.5} & Cond & 28 & 7.2 & 5.94\\
& CFG & 32 & 15.7 & 6.01\\
& DG & 32 & 10.6 & 6.16\\
\midrule
\multirow{3}{*}{Flux1} & Cond & 35 & 16.2 & 5.50\\
& CFG & 42 & 36.0 & 5.96\\
& DG & 42 & 24.8 & 6.13\\
\bottomrule
\end{tabular}}
\vspace{-3mm}
\end{table}

\subsection{Ablation Studies}
\subsubsection{Core components in DG.}
We ablate the core design choices of DG on SD3 conditional generation. As shown
in~\Cref{tab:ablation_dg}, simply constructing a guidance branch from random
channels brings only limited gains, indicating that the selected MA dimensions
are important for isolating detail-related changes. Replacing the adaptive
timestep schedule with a fixed suppression coefficient improves over the
baseline, but remains weaker than the full design. This suggests that preserving
MAs during early structure formation and suppressing them more strongly in later
denoising stages is beneficial. The full DG variant, which combines calibrated
MA dimension selection, adaptive MA suppression, and residual detail guidance,
achieves the best detail-quality metrics.

\begin{table}[t]
\setlength\tabcolsep{1pt}
\centering
\caption{\textcolor{black}{\textbf{Evaluation of DG on non-standard DiTs.} 
Our DG strategy generalizes effectively to other architectures and significantly 
enhances the visual detail quality for both Pixart-Alpha and SANA.}}
\label{tab:non_dit}
\resizebox{\linewidth}{!}{
\begin{tabular}{ccccccc}
\toprule
\multirow{2}{*}{Model} & \multirow{2}{*}{Type} & \multirow{2}{*}{DG} & \multicolumn{2}{c}{Prompt Alignment} & \multicolumn{2}{c}{Detail Quality} \\
\cmidrule(lr){4-5}\cmidrule(lr){6-7}
& &  & Clipscore & Blipscore & HPSv2.1 & Aesthetic \\
\midrule
\multirow{4}{*}{PixArt-alpha}& \multirow{2}{*}{Cond}& $\times$ & 22.64 & 68.41 & 25.63 & 6.01\\
&& $\checkmark$ & \textbf{23.43} & \textbf{72.07} & \textbf{29.18} & \textbf{6.53} \\
\cmidrule(lr){2-7}
&\multirow{2}{*}{CFG}& $\times$ & \textbf{26.20} & \textbf{87.64} & 29.99 & 6.21\\
&& $\checkmark$ & 26.17 & 86.88 & \textbf{30.74} & \textbf{6.34} \\
\midrule
\multirow{4}{*}{SANA}& \multirow{2}{*}{Cond}& $\times$ & 23.52 & 78.25 & 24.40 & 5.91\\
&& $\checkmark$ & \textbf{24.98} & \textbf{84.11} & \textbf{28.72} & \textbf{6.12} \\
\cmidrule(lr){2-7}
&\multirow{2}{*}{CFG}& $\times$ & 27.07 & \textbf{91.03} & 30.13 & 6.00\\
&& $\checkmark$ & \textbf{27.20} & 90.25 & \textbf{30.52} & \textbf{6.07} \\
\bottomrule
\end{tabular}}
\vspace{-3mm}
\end{table}


\subsubsection{Core components in MREP.}
We further ablate the representation-side design of MREP. The method contains
two complementary components: AdaLN-based modulation reduces the directional
dominance of MA channels, while normalized MA spatial maps preserve useful
spatial response patterns. As reported in~\Cref{tab:ablation_mrep}, using AdaLN
modulation alone already improves dense prediction over raw DiT features, showing
that suppressing MA dominance is important. Adding MA maps without spatial
normalization is less stable, because their large magnitudes can again dominate
feature similarity. The full MREP representation combines AdaLN-modulated
features with normalized MA maps and obtains the best overall results across
correspondence, segmentation, and depth estimation.

\begin{figure*}[t]
  \centering
  \includegraphics[width =\linewidth]{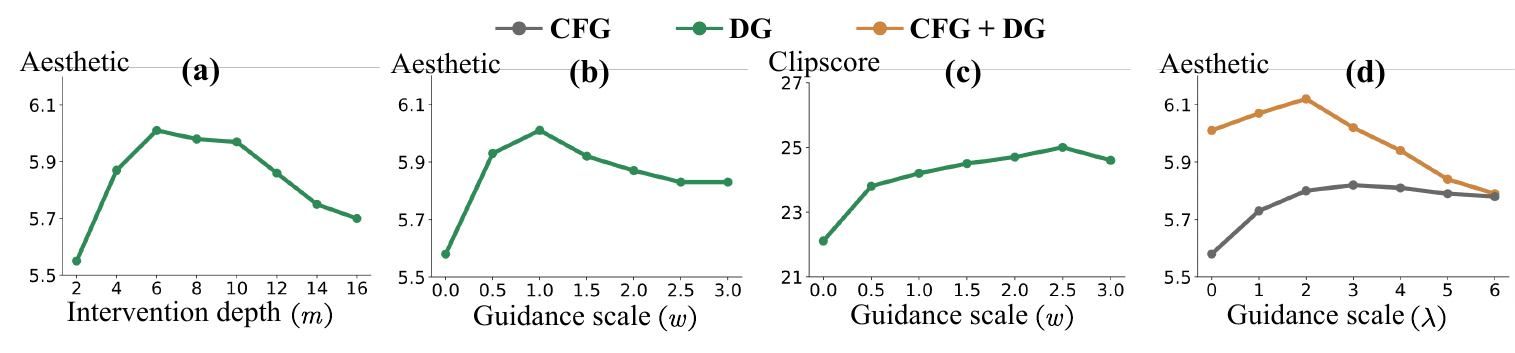}
  \caption{\textbf{Investigations of intervention depth $m$, scales $\lambda$ and $w$ for
  SD3.} We analyze how the DG intervention block, CFG scale $\lambda$, and DG
  scale $w$ affect generation quality. The results show that DG benefits from
  intervening at appropriate intermediate blocks and provides a flexible
  trade-off between semantic alignment and fine-grained detail enhancement.}
  \vspace{-2mm}
  \label{figure:f8}
\end{figure*}
\begin{figure}[t]
  \centering
  \includegraphics[width=\linewidth]{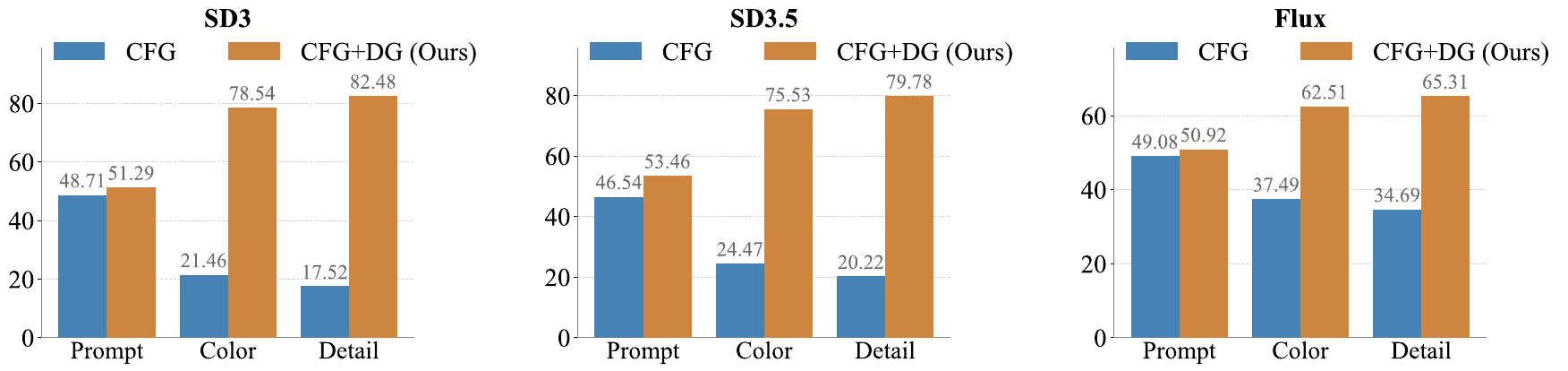}
  \caption{\textbf{User study on SD3, SD3.5, and Flux1.} We report win rates comparing
  CFG with CFG+DG from prompt alignment, detail preservation, and color consistency.}
  \vspace{-3mm}
  \label{figure:user_study}
\end{figure}

\begin{table}[t]
\centering
\caption{\textbf{Ablation of core DG components on SD3.} We report HPSv2.1 and
Aesthetic scores under conditional generation.}
\label{tab:ablation_dg}
\normalsize
\setlength\tabcolsep{2pt}
\renewcommand{\arraystretch}{1.08}
\begin{tabular*}{\linewidth}{@{}l@{\extracolsep{\fill}}cccc@{}}
\toprule
Variant & MA & $\rho_t$ & HPSv2.1 & AES \\
\midrule
Cond baseline & $\times$  & $\times$ & 21.84 & 5.58 \\
Random channel & $\times$ & $\checkmark$  & 23.42 & 5.69 \\
w/o adaptive schedule & $\checkmark$ & $\times$  & 27.16 & 5.88 \\
Full DG & $\checkmark$ & $\checkmark$& \textbf{28.65} & \textbf{6.01} \\
\bottomrule
\end{tabular*}
\vspace{-3mm}
\end{table}

\begin{table}[t]
\centering
\caption{\textbf{Ablation of core MREP components.} We report representative metrics
on semantic correspondence, segmentation, and depth estimation.}
\label{tab:ablation_mrep}
\normalsize
\setlength\tabcolsep{2pt}
\renewcommand{\arraystretch}{1.08}
\begin{tabular*}{\linewidth}{@{}l@{\extracolsep{\fill}}ccc@{}}
\toprule
Variant & SPair$\uparrow$ & ADE$\uparrow$ & NYUv2$\downarrow$ \\
\midrule
Raw DiTF & 64.0 & 54.8 & 0.256 \\
AdaLN modulation only & 64.4 & 55.5 & 0.231 \\
MA maps w/o norm. & 63.8 & 55.2 & 0.239 \\
Full MREP & \textbf{64.7} & \textbf{56.1} & \textbf{0.220} \\
\bottomrule
\end{tabular*}
\vspace{-3mm}
\end{table}

\subsubsection{Intervention depth.} 
We examine the effect of the intervention depth $k=m$
of massive activations, as shown in~\Cref{figure:f8}(a). Our DG strategy achieves the
best performance when applied to intermediate blocks (e.g., $k$ ranging from 4 to 10).
We hypothesize that early blocks mainly contain heavy noise and lack even coarse image
structures, making disruption there uninformative, while applying disruption in late
blocks occurs too close to the final output and thus has minor impact on generation.
Based on these observations, we primarily perturb massive activations in the
intermediate layers and set the default $k=6$ for the SD3 model. For SD3.5 and Flux, we
use $k=20$ and $k=22$, respectively, following their layer-wise MA distributions.
\subsubsection{Guidance scales.} We present the quantitative
results across different scales in~\Cref{figure:f8}. Our DG consistently achieves
stable and high Aesthetic (AES) scores (\Cref{figure:f8}(b)) and CLIPScore
(\Cref{figure:f8}(c)). When combined with CFG, it further boosts AES performance
(\Cref{figure:f8}(d)). These results highlight the effectiveness and robustness of our
approach in enhancing fine-grained details. Moreover, DG integrates seamlessly with
CFG, enabling joint improvements in local-detail fidelity and prompt alignment.
\subsubsection{User study.} We conduct a user study to evaluate the benefits of our
DG strategy from three key aspects: prompt alignment, color consistency, and detail
preservation. For each model, 20 annotators compare 100 image pairs generated by CFG
and CFG+DG. As shown in~\Cref{figure:user_study}, CFG+DG obtains higher preference
rates, especially on detail preservation and color consistency, demonstrating that the
automated detail-quality gains are aligned with human judgment.

\section{Conclusion}
In this paper, we presented a systematic study of Massive Activations in
Diffusion Transformers and showed that they form a distinctive activation
structure for dense visual computation: they are spatially distributed across
image tokens, concentrated in a small set of feature dimensions, closely related
to AdaLN modulation, and primarily governed by the denoising stage. This
structure gives MAs a dual role. For generation, they are crucial for
fine-grained local detail synthesis while having limited impact on global
semantics. For understanding, their shared high-magnitude directions can dominate
raw DiT features and weaken spatial discrimination.

Based on these observations, we introduced EMA, a training-free framework that
elicits the capabilities of pretrained DiTs through task-aware MA modulation. For
visual generation, DG constructs a detail-deficient counterfactual branch by
suppressing MA dimensions and uses it to guide sampling toward more detailed
outputs. DG is compatible with CFG, supports partial-forward inference, and can
be applied locally to selected spatial regions. For visual understanding, MREP
reduces MA directional dominance with AdaLN-based modulation while preserving
spatial MA cues, producing more discriminative dense representations.
Experiments across image generation, video generation, local detail refinement,
semantic correspondence, semantic segmentation, and depth estimation demonstrate
that EMA consistently improves both the generative quality and representation
utility of DiTs. We hope this work encourages further exploration of internal
activation structures as controllable and reusable signals in large visual
generative models.

\bibliographystyle{IEEEtran}
\bibliography{main}

\begin{IEEEbiographynophoto}{Chaofan Gan}
received the Bachelor's degree in Software Engineering from Huazhong University of
Science and Technology, Wuhan, China in 2022. He is currently pursuing the Ph.D. degree
in Electronic Information and Electrical Engineering at Shanghai Jiao Tong University,
Shanghai, China. His research interests include video generation and understanding,
diffusion models, and noise learning.
\end{IEEEbiographynophoto}
\begin{IEEEbiographynophoto}{Zicheng Zhao} received the Bachelor's degree in Information Engineering from Shanghai Jiao Tong University, Shanghai, China, in 2025. He is currently pursuing the Master's degree in Information Engineering at Shanghai Jiao Tong University, Shanghai, China. His research interests include video large language models, video understanding, and diffusion models.
\end{IEEEbiographynophoto}
\begin{IEEEbiographynophoto}{Yuanpeng Tu}
received the Master and Bachelor degree from Tongji University in 2020 and 2023. He is
currently pursuing the Ph.D. degree in Computer Science at The University of Hong Kong,
Hong Kong, China. His current research mainly focuses on video/image generation, robust
learning (i.e., noisy labels), open-world/zero-shot perception, and label-efficient
learning.
\end{IEEEbiographynophoto}
\begin{IEEEbiographynophoto}{Xi Chen}
received the BE and MS from Zhejiang University in 2017 and 2020, and received the
double MS degree from Ecole Centrale de Marseille (France) in 2020. He is currently
working towards the PhD degree with the Department of Computer Science, The University
of Hong Kong. His current research interests include computer vision and content
generation, especially in image/video generation and editing.
\end{IEEEbiographynophoto}
\begin{IEEEbiographynophoto}{Ziran Qin} received the M.E. degree from Shanghai Jiao Tong University in
2023. He is currently pursuing the Ph.D. degree with the School of Electronic
Information and Electrical Engineering, Shanghai Jiao Tong University. His research
interests include efficient generative modeling, visual autoregressive generation, and
model compression.\end{IEEEbiographynophoto}
\begin{IEEEbiographynophoto}{Tieyuan Chen} 
received the Bachelor’s degree in the school of electronic information and electrical
engineering from Sichuan University in 2023. He is currently pursuing the Ph.D. degree
at Shanghai Jiao Tong University and the Zhongguancun Academy. His research interests
include causal discovery, causal reasoning, and video reasoning.
\end{IEEEbiographynophoto}
\begin{IEEEbiographynophoto}{Supavadee Aramvith} (S’95) received the B.S.
(first-class honors) degree in computer science from Mahidol University, Bangkok,
Thailand, in 1993, and the M.S. degree in electrical engineering from University of
Washington at Seattle in 1996, where she is working toward the Ph.D. degree under the
scholarship of The Royal Thai Government. Her research interests include rate-control,
rate-distortion optimization video coding, and transmission of video over wireless
channels and heterogeneous networks.\end{IEEEbiographynophoto}
\begin{IEEEbiographynophoto}{Junhui Hou} (Senior Member, IEEE) received the
BEng degree in information engineering (Talented Students Program) from the South China
University of Technology, Guangzhou, China, in 2009, the MEng degree in signal and
information processing from Northwestern Polytechnical University, Xian, China, in
2012, and the PhD degree in electrical and electronic engineering from the School of
Electrical and Electronic Engineering, Nanyang Technological University, Singapore, in
2016. In January 2017, he joined the Department of Computer Science, City University of
Hong Kong, as an assistant professor. His research focuses on visual computing, such as
image or video or 3D geometry data representation, processing and analysis, semi or
unsupervised data modeling, and data compression and adaptive transmission.\end{IEEEbiographynophoto}
\begin{IEEEbiographynophoto}{Mehrtash Harandi} received the BSc in Electronics from Sharif University of technology, MSc and PhD 
degrees in Computer Science from the University of Tehran, Iran. Dr. Harandi is a
senior researcher at Computer Vision Research Group (CVRG), NICTA. His main research
interests are theoretical and computational methods in computer vision and machine
learning with a focus on Riemannian geometry.
\end{IEEEbiographynophoto}
\begin{IEEEbiographynophoto}{Weiyao Lin}
(Senior Member, IEEE) received the B.E. and M.E. degrees from Shanghai Jiao Tong
University, Shanghai, China, in 2003 and 2005, respectively, and the Ph.D. degree from
the University of Washington, Seattle, WA, USA, in 2010, all in electrical engineering.
He is currently a Professor at the Department of Electronic Engineering, Shanghai Jiao
Tong University. He has authored or coauthored more than 100 technical articles on top
journals/conferences including the IEEE TRANSACTIONS ON PATTERN ANALYSIS AND MACHINE
INTELLIGENCE, the International Journal of Computer Vision, the IEEE TRANSACTIONS ON
IMAGE PROCESSING, CVPR, NeurIPS, ICLR, and ICCV. He holds more than 20 patents. His
research interests include video/image analysis, computer vision, and video/image
processing applications.
\end{IEEEbiographynophoto}

\end{document}